\documentclass{article}

\PassOptionsToPackage{numbers,sort&compress}{natbib}
\usepackage[preprint]{neurips_2026}
\usepackage[utf8]{inputenc} % allow utf-8 input
\usepackage[T1]{fontenc}    % use 8-bit T1 fonts
\usepackage{hyperref}       % hyperlinks
\usepackage{url}            % simple URL typesetting
\usepackage{booktabs}       % professional-quality tables
\usepackage{amsfonts}       % blackboard math symbols
\usepackage{nicefrac}       % compact symbols for 1/2, etc.
\usepackage{microtype}      % microtypography
\usepackage{multirow}
\usepackage{graphicx}
\usepackage{amsmath}
\usepackage{amssymb}
\usepackage{amsthm}
\usepackage{subcaption}
\usepackage[table]{xcolor}
\newtheorem{definition}{Definition}
\def\ie{\textit{i.e.}} % that is
\newcommand{\best}[1]{\textbf{#1}}
\newcommand{\second}[1]{\underline{#1}}
\newcommand{\cmark}{\checkmark}

\title{Robust Multi-view Clustering against \\ Imperfect Information}

\author{%
  Zhichao Huang$^1$ \quad
  Haochen Zhou$^1$ \quad
  Hao Wang$^1$ \quad
  Mouxing Yang$^{1}$ \quad
  Xi Peng$^{2,3}$ \\[1ex]
  $^1$College of Computer Science, Sichuan University, China. \\
  $^2$School of Artificial Intelligence, Sichuan University, China. \\
  $^3$Tianfu Jincheng Laboratory, Chengdu, China. \\[1ex]
  \texttt{\{zhichaohuang.gm, haochenzhou.xl, cshaowang, yangmouxing, pengx.gm\}@gmail.com}
}

\begin{document}

\maketitle

\begin{abstract}
  Real-world multi-view data always suffer from imperfect information problem, where the view-specific observations are absent (\ie, Incomplete Views, IV) and cross-view correspondences are mismatched (\ie, Noisy Correspondences, NC) for certain instances.
  As a remedy, numerous IV- and NC-oriented multi-view clustering (MvC) methods have been proposed, which however require either reliable correspondences or sufficiently complete instances, thus stopping short of addressing the imperfect information problem.
  In contrast, we observe that both IV and NC challenges originate from the same issue of imperfect cross-view counterpart information, where the counterpart of an anchor instance in another view might be either unavailable or unreliable.
  Based on the observation, we propose a novel robust MvC framework, termed Posterior-guided Latent Counterpart Inference (PLCI), which could handle both IV and NC in a unified manner.
  Specifically, PLCI formulates the desired cross-view counterpart of each anchor instance as a latent variable, and integrates both instance-level reliability and prototype-level semantic transport to infer the posterior distribution of the latent counterpart. 
  Extensive experiments on six widely-used multi-view datasets against 10 state-of-the-art MvC methods demonstrate the effectiveness of PLCI for tackling the imperfect information problem.
  The code will be released upon acceptance.
\end{abstract}

\section{Introduction}
\begin{figure}
  \centering
  \includegraphics[width=\linewidth]{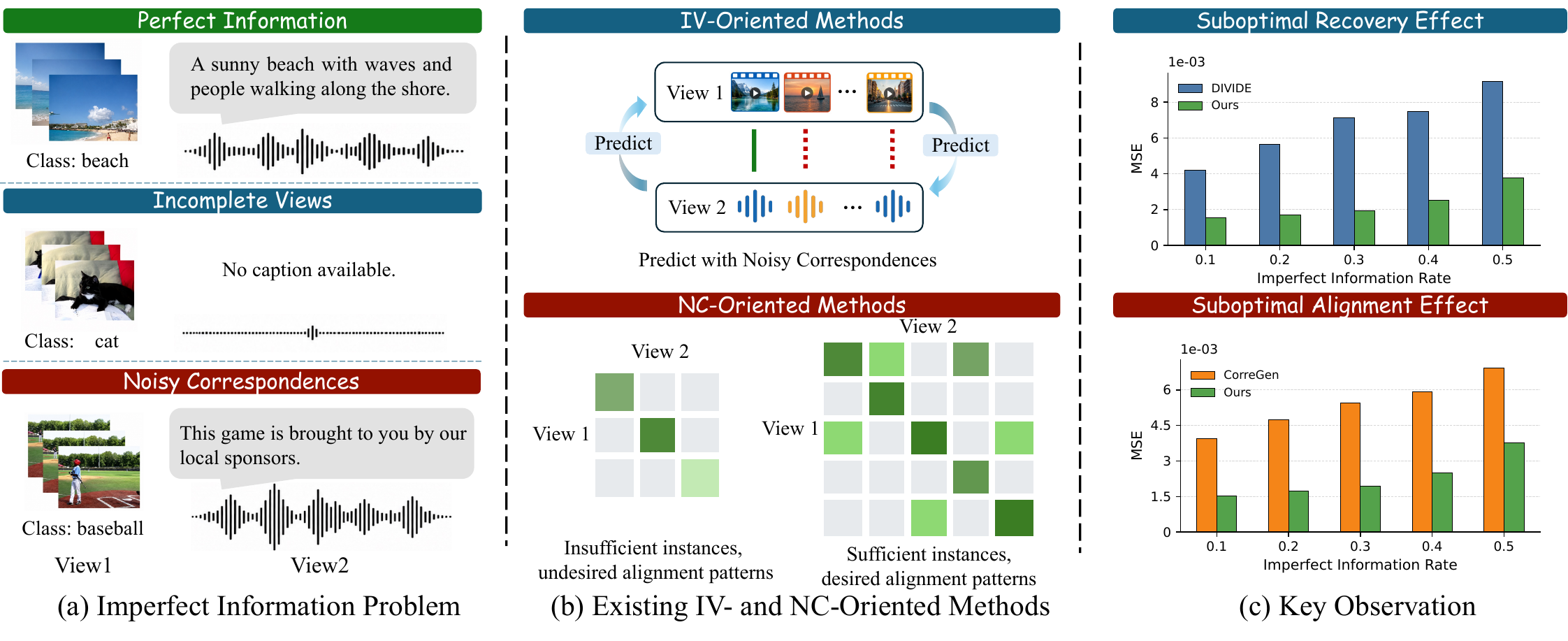}
  \caption{Our motivations and observations.
  (a) The studied problem. Ideally, the video and audio views should provide consistent semantics. However, in the noisy correspondences (NC) case, the audio view is irrelevant to the visual content. In the incomplete views (IV) case, the audio view is missing.
  (b) Limitations of existing IV- and NC-oriented methods. Under imperfect information, IV-oriented methods are susceptible to unreliable correspondences during training, which could result in erroneous view recovery. Meanwhile, NC-oriented methods would establish inferior alignment effects across views as the number of complete instances for correspondence inference is insufficient.
  (c) Analytic studies. With the increasing imperfect information level, the recovery/alignment effect of both state-of-the-art IV- and NC-oriented methods significantly degrades, demonstrating that neither IV- nor NC-oriented methods can handle the imperfect information problem. The analyses are conducted on the Scene15 dataset, and the recovery/alignment effect is measured by the mean squared error (MSE) between the recovered/aligned view and the ground-truth view.}
  \label{fig1}
\end{figure}

Real-world instances are always characterized by multiple views or modalities, including but not limited to RGB images, depth maps, audio signals, and text descriptions. 
As one of the most fundamental tools for analyzing such heterogeneous data, multi-view clustering (MvC) aims to partition unlabeled instances into semantic groups by exploiting complementary information within individual views and consistency information across cross-view correspondence~\cite{du2025openviewer,fang2024representation,liu2021efficient,liu2021onepass,wang2019multiview,wang2022highly,qin2024dual}. 
In practice, however, the success of most existing MvC methods is often hindered by the imperfect information problem, which is mainly manifested as incomplete views (IV) and noisy correspondences (NC).
Specifically, as illustrated in Fig. \ref{fig1}(a), IV and NC refer to the absence of view-specific observations and the mismatched inter-view correspondences for certain instances, respectively.
As a result, these two challenges would impede the exploitation of both view-specific and cross-view information, thereby significantly degrading MvC performance.

To conquer these challenges, numerous robust MvC methods~\cite{liu2021onepasslarge,liu2023contrastive,liang2025concrete,liu2024learn} have been proposed.
Specifically, as shown in Fig.~\ref{fig1}(b)-(c), for the IV challenge, IV-oriented MvC methods typically learn the cross-view generative mapping from the complete instances, and accordingly recover the missing views of incomplete instances with the help of the observed ones.
However, mismatched cross-view correspondences might steer the optimization of generators toward incorrect cross-view mappings, resulting in suboptimal or even erroneous recovery.
Different from the widely-studied IV challenge, the NC challenge has attracted increasing attention in recent years.
In general, NC-oriented methods aim to re-establish reliable inter-view correspondence by leveraging the alignment pattern learned from a number of complete instances.
Despite their robustness against NC, the lack of sufficient complete instances would undermine the learning of reliable cross-view alignment patterns.
Clearly, existing studies~\cite{wen2018incomplete,wen2020dimc,yang2023dealmvc,yang2025generalized} are mainly designed to handle either the IV or NC challenge.
In other words, how to address both challenges in a unified manner and thus tackle the imperfect information problem remains under-explored.

In this paper, we observe that both IV and NC originate from imperfect cross-view counterpart information.
Specifically, IV makes the counterpart unavailable, whereas NC renders the observed counterpart unreliable.
One straightforward way to address these challenges is to combine the existing solutions of IV and NC.
To be more specific, one could first recover the unavailable counterparts and then identify the reliable ones in other views, or vice versa. 
However, such a two-stage strategy is inherently suboptimal, as errors introduced in either recovery or re-alignment would inevitably propagate to the other stage. 
More importantly, as aforementioned, the effectiveness of view recovery and correspondence re-alignment fundamentally depends on reliable correspondences and sufficiently complete instances, both of which are daunting to satisfy in the context of imperfect information.
To support the above claims, we present analysis experiments in Table~\ref{tab:combine_order}.

To solve the imperfect information problem, we propose a novel MvC method termed Posterior-guided Latent Counterpart Inference (PLCI), which could address both IV and NC challenges in a unified framework.
The key idea behind PLCI is reformulating the counterpart of instances as a latent variable and inducing the posterior distribution for the latent counterpart by integrating both instance-level reliability and prototype-level semantic transport.
Specifically, PLCI first estimates the correspondence reliability of complete instances by resorting to the memory effect of deep neural networks~\cite{dnnmemorization}.
After that, PLCI performs reliability-aware optimal transport upon view-specific prototypes constructed from all observed samples, thus propagating semantics across views.
Finally, by combining the transported semantic structure with the observed information, PLCI could infer the posterior distribution for each instance and derive the reliable pseudo counterpart for achieving robust multi-view clustering.
In brief, the main contributions of this work are outlined as follows:
\begin{itemize}
  \item We study a practical yet under-explored problem in multi-view clustering, termed imperfect information, where incomplete views and noisy correspondences simultaneously exist. Moreover, we investigate the negative impact of the imperfect information problem and reveal that existing IV- and NC-oriented methods are incapable of handling the problem.
  \item To solve the imperfect information problem, we propose a novel MvC method, termed Posterior-guided Latent Counterpart Inference (PLCI). Instead of handling unavailable or unreliable counterparts separately, PLCI formulates the counterpart of each instance as a latent variable, thus addressing the imperfect information problem in a unified framework. 
  \item Extensive comparative experiments against 10 state-of-the-art robust MvC methods on six widely-used multi-view clustering datasets covering diverse modalities (\textit{e.g.}, video-audio, RGB-depth), verify the effectiveness of PLCI. Furthermore, we demonstrate that PLCI is a generalizable framework that could be seamlessly incorporated into a broad spectrum of existing MvC methods to enhance their robustness against imperfect information.
\end{itemize}

\section{Method}
\label{Method}
\begin{figure}
  \centering
  \includegraphics[width=\linewidth]{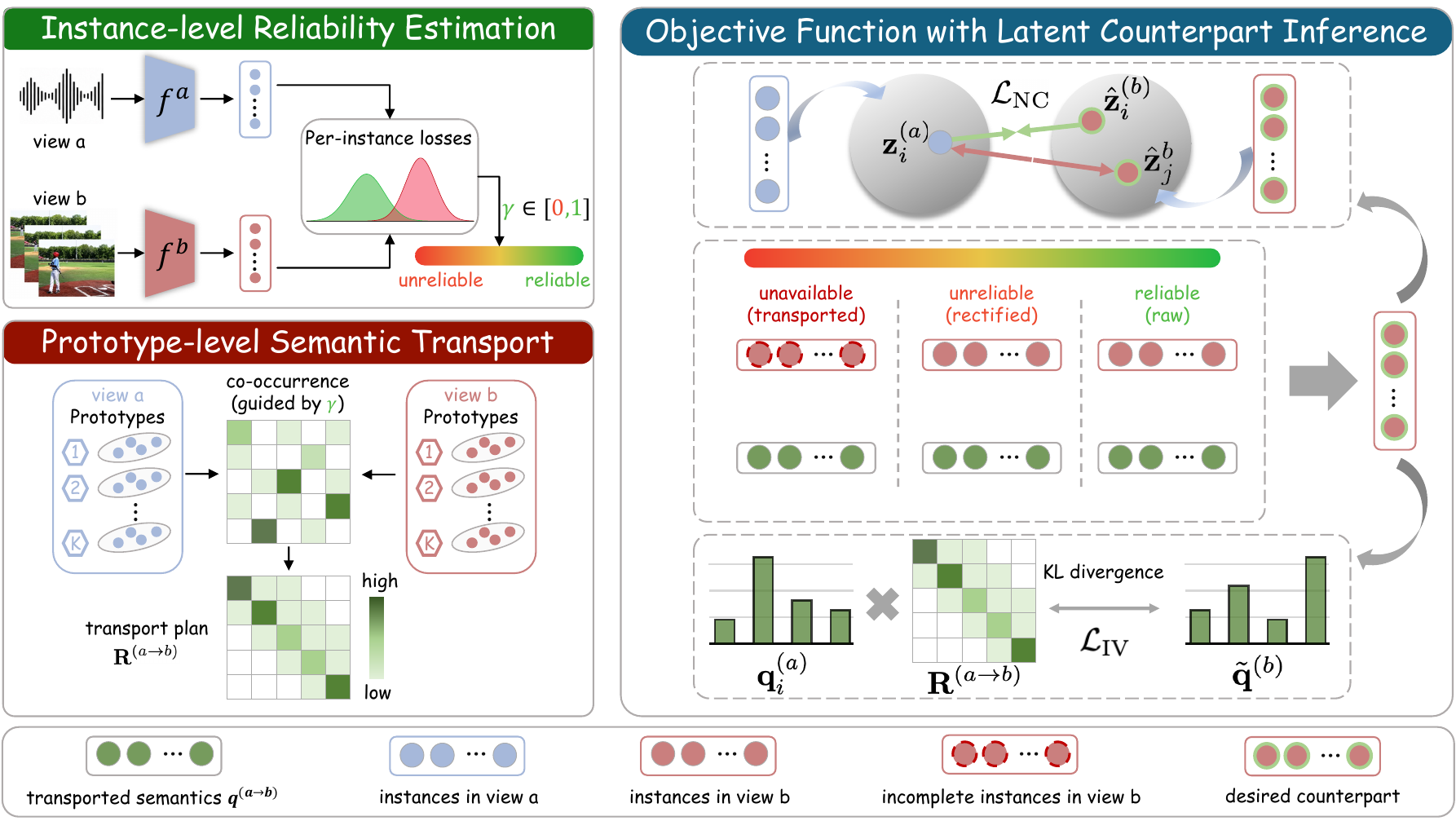} 
  \caption{\textbf{Overview of PLCI.}
  To begin with, PLCI estimates the instance-level reliability $\gamma_i$ from the pair-loss distribution of complete instances. 
  Then, PLCI constructs view-specific prototypes and computes a reliability-aware co-occurrence matrix, which serves as a robust proxy for cross-view prototype correspondence and thus facilitates prototype-level semantic transport. 
  Finally, PLCI integrates the observed counterpart information and the transported semantic counterpart to infer the latent counterpart for achieving robust multi-view clustering against imperfect information.
  }
  \label{fig2}
\end{figure}

In this section, we present {Posterior-guided Latent Counterpart Inference (PLCI)} for robust multi-view clustering under imperfect information. 
As shown in Fig.~\ref{fig2}, PLCI formulates the desirable target-view counterpart as a latent variable, and infers it by combining instance-level correspondence reliability with prototype-level semantic transport. 
Specifically, Sec.~\ref{subsec:problem} defines the imperfect information problem and the latent counterpart formulation. 
Sec.~\ref{subsec:reliability} estimates reliable correspondences from complete instances, while Sec.~\ref{subsec:semantic_transport} propagates semantic structures across views via prototype-level transport. 
Finally, Sec.~\ref{subsec:objective} presents the latent counterpart inference criterion and the joint objective for learning with both noisy correspondences and incomplete views.

\subsection{Problem Formulation}
\label{subsec:problem}

Given a multi-view dataset $\mathcal{D}=\{\mathcal{X}_1, \mathcal{X}_2, \cdots, \mathcal{X}_N\}$, the objective of MvC is to construct a common space where $N$ instances could be partitioned into $C$ groups with different semantics. 
For clarity, let $\mathcal{X}_{i}=\{(\mathbf{x}_i^{(v)} \mid m_i^{(v)}\in\{0,1\})\}_{v=1}^{V}$ denote the $i$-th instance,
where $v$ is the view index and $m_i^{(v)}=0/1$ indicates that the view $v$ is unavailable/available. 
Besides, in the following, we use $a$ and $b$ ($b\neq a$) to denote the view indices of the anchor and target views, respectively. 
To begin with, we define the imperfect information problem as follows.
\begin{definition}[Imperfect Information]
\label{def:imperfect_information}
The dataset $\mathcal{D}$ is contaminated with imperfect information when there exists a subset $\mathcal{D}_{\mathrm{imp}}$ that satisfies the following condition:
\begin{equation}
\begin{aligned}
\mathcal{D}_{\mathrm{imp}}
&=
\Biggl\{
\mathcal{X}_i \ \Biggm|\
\underbrace{0<\sum_{v=1}^{V} m_i^{(v)} < V}_{\mathrm{incomplete\ views}}
\ \vee\ 
\underbrace{
\sum_{\substack{b=1, b\neq a}}^{V}
\mathbb{I}\!\left(\mathbf{x}_i^{(b)}, \mathbf{x}_i^{(a)}\right)<(V-1)
}_{\mathrm{noisy\ correspondences}}
\Biggr\},
\end{aligned}
\label{eq:imperfect_set}
\end{equation}
where $\mathbb{I}(\cdot, \cdot)$ is the indicator function evaluating to 1 i.f.f. $\mathbf{x}_i^{(b)}$ and $\mathbf{x}_i^{(a)}$ belong to the same instance.
\end{definition}

Definition~\ref{def:imperfect_information} demonstrates that imperfect information is mainly manifested as Incomplete Views (IV) and Noisy Correspondences (NC).
To be specific, for an observed anchor $\mathbf{x}_i^{(a)}$, IV contributes to the absence of its counterparts in other views, while NC results in the presence of unreliable counterparts, both of which would heavily degrade the MvC performance.
However, most existing IV-oriented and NC-oriented MvC methods are capable of handling either one of these two challenges, stopping short of addressing the imperfect information problem, as verified in Table~\ref{tab:main_results} and Figs.~\ref{fig:scene15_fixed_fp}-\ref{fig:scene15_fixed_mr}.  

To tackle the imperfect information problem, one can formulate the desirable counterpart representation in the target view as a latent variable $\hat{\mathbf{z}}_i^{(b)}$ and model its posterior distribution as 
\begin{equation}
p_{\phi}
\left(
\hat{\mathbf{z}}_i^{(b)}
\mid
\mathcal{O}_i
\right),
\label{eq:latent_counterpart}
\end{equation}
where $\mathcal{O}_i$ denotes the combination of $i$-th instance observations and the whole dataset $\mathcal{D}$ acting as prior for the latent counterpart $\hat{\mathbf{z}}_i^{(b)}$.
Ideally, Eq.~\ref{eq:latent_counterpart} should shed light on the information inherent in $\mathcal{O}_i$, so that one could obtain the desirable counterpart for each anchor instance.
However, the potential noisy correspondences in the complete instances and the absence of counterparts in incomplete instances would prevent the effective inference of the latent counterparts.

As a remedy, we propose PLCI to enable effective latent counterpart inference for robust multi-view clustering under imperfect information. 
PLCI consists of two modules (instance-level reliability estimation and prototype-level semantic transport) to explore and exploit information in complete instances and incomplete instances for counterpart inference, and a joint objective function to achieve robust multi-view clustering against imperfect information.

\subsection{Instance-level Reliability Estimation}
\label{subsec:reliability}
The obstacle behind the information mining for complete instances is the potential noisy correspondences, which would mislead the cross-view common space construction.
As a remedy, the proposed method aims to estimate the correspondence reliability of complete instances, so that the NC-contaminated complete instances could be uncovered.
To this end, the proposed method resorts to the widely-acknowledged memorization effect of deep neural networks (DNNs)~\cite{dnnmemorization}, which indicates that DNNs tend to fit simple patterns and gradually learn harder ones, thus leading to relatively lower losses for clean samples.
Motivated by this, one can regard the loss distribution of complete instances as a proxy for estimating the reliability of their correspondence.
To be specific, for each complete instance, the per-instance losses are computed as follows,
\begin{equation}
\left\{
\ell_{\mathrm{con}}
\left(
f^{(a)}(\mathbf{x}_i^{(a)}),
f^{(b)}(\mathbf{x}_i^{(b)})
\right)
\ \middle|\
m_i^{(a)}m_i^{(b)}=1
\right\},
\label{eq:pair_loss_set}
\end{equation}
where $f^{(v)}(\cdot)$ denotes the view-specific encoder for view $v$ and $\ell_{\mathrm{con}}(\cdot, \cdot)$ is the loss function for measuring the consistency between two views.
In the implementation, one can adopt the widely-used contrastive loss~\cite{divide,simclr,moco} as $\ell_{\mathrm{con}}(\cdot, \cdot)$.

After that, following the typical paradigm in noisy label learning~\cite{dividemix}, one could fit the loss value set in Eq.~\ref{eq:pair_loss_set} with a two-component Gaussian Mixture Model (GMM).
Formally,
\begin{equation}
  p(\ell_{\mathrm{con}} \mid \theta) = \sum_{k=1}^{2} \pi_k \Psi(\ell_{\mathrm{con}} \mid k),
\end{equation}
where $\pi_k$ and $\Psi(\ell_{\mathrm{con}} \mid k)$ denote the mixture coefficient and the Gaussian distribution of the $k$-th component, respectively.
According to the memory effect of DNNs, the component with a smaller mean is expected to represent instances with reliable correspondence.
Therefore, one could estimate the correspondence reliability $\gamma_i$ for complete instances by the posterior probability of their losses belonging to the low-loss component $\kappa$, namely,
\begin{equation}
  \gamma_i = p(\kappa \mid \ell_{\mathrm{con}}).
\end{equation}

\subsection{Prototype-level Semantic Transport}
\label{subsec:semantic_transport}
To mine the information from incomplete instances, it is expected to complement the missing views with the help of observed ones.
Unlike the existing IV-oriented methods that directly recover the missing views at the instance level, PLCI transports the semantic structure across views.
The motivation behind this design is two-fold. First, the recovery effect of existing methods heavily relies on sufficient complete instances with reliable correspondences, which is hard to satisfy in the context of imperfect information.
In contrast, the semantic structure of each view can be estimated from all available observations, thus being more robust against imperfect information.
Second, the propagated semantics across views encourage cross-view consistency, which could further facilitate the cross-view common space construction.

Specifically, PLCI first reveals the semantic structure of each view by maintaining a set of view-specific prototypes.
Formally speaking, for each view $v$, PLCI maintains $C$ prototypes 
$\{\boldsymbol{\mu}_c^{(v)}\}_{c=1}^{C}$.
After that, PLCI estimates the semantic structure of each view by assigning each observed instance to the prototypes.
Formally, for an observed sample $\mathbf{x}_i^{(v)}$, the semantic structure is defined as 
\begin{equation}
q_{ic}^{(v)}
=
\frac{
\exp\left(\mathrm{sim}(\mathbf{z}_i^{(v)},\boldsymbol{\mu}_c^{(v)})/\tau_p\right)
}{
\sum_{r=1}^{C}
\exp\left(\mathrm{sim}(\mathbf{z}_i^{(v)},\boldsymbol{\mu}_r^{(v)})/\tau_p\right)
},
\label{eq:prototype_assignment}
\end{equation}
where $\mathbf{z}_i^{(v)}=f^{(v)}(\mathbf{x}_i^{(v)})$, $\tau_p$ is the temperature parameter and $\mathrm{sim}(\cdot,\cdot)$ denotes the similarity function. 
Therefore, the semantic structure of $\mathbf{x}_i^{(v)}$ is characterized by the assignment vector $\mathbf{q}_i^{(v)}=[q_{i1}^{(v)},\cdots,q_{iC}^{(v)}]$. 

To propagate semantic structure across views, PLCI establishes cross-view prototype correspondence by employing reliability-aware optimal transport. 
To be specific, PLCI estimates the co-occurrence matrix of prototypes across views, which could reflect the likelihood of different cross-view prototypes describing the same underlying semantics.
However, the estimation of co-occurrence matrix is susceptible to noisy correspondences.
To mitigate the adverse effect of noisy correspondences, the estimated reliability of complete instances is further incorporated into the co-occurrence estimation.
Mathematically, the co-occurrence matrix is defined as 
\begin{equation}
\mathbf{U}^{(a,b)}
=
\sum
\gamma_i
\left(\mathbf{q}_i^{(a)}\right)^{\top}
\mathbf{q}_i^{(b)},
\label{eq:co_matrix}
\end{equation}
where $\gamma_i$ denotes the estimated reliability of the correspondence between $\mathbf{x}_i^{(a)}$ and $\mathbf{x}_i^{(b)}$. 
Clearly, a larger value of $\mathbf{U}_{tr}^{(a,b)}$ indicates that the $t$-th prototype in view $a$ and the $r$-th prototype in view $b$ are more likely to describe the same underlying semantics.

After that, one can employ the optimal transport to derive the cross-view prototype correspondence, which indicates how the semantic structure of one view should be transported to the other view.
To be specific, the cost matrix for the optimal transport is defined as
\begin{equation}
\mathbf{D}^{(a,b)}
=
-\frac{\mathbf{U}^{(a,b)}}{\|\mathbf{U}^{(a,b)}\|_{\infty}},
\label{eq:transport_cost}
\end{equation}where $\|\cdot\|_{\infty}$ denotes the infinity norm.
After solving the optimal transport problem with the cost matrix $\mathbf{D}^{(a,b)}$, one could obtain the transport plan $\mathbf{R}^{(a\rightarrow b)}$.
Finally, the prototype-level semantic transport from view $a$ to view $b$ could be achieved as follows,
\begin{equation}
\mathbf{q}_i^{(a\rightarrow b)}
=
\mathbf{q}_i^{(a)}
\mathbf{R}^{(a\rightarrow b)}.
\label{eq:transported_assignment}
\end{equation}

\subsection{Objective Function with Latent Counterpart Inference}
\label{subsec:objective}

After obtaining the instance-level correspondence reliability and the prototype-level semantic transport, PLCI could infer the latent counterpart by combining the observed counterpart information with the transported semantic prior. 
Specifically, one could use the posterior mean as the inferred counterpart, which is approximated as
\begin{equation}
\hat{\mathbf{z}}_i^{(b)}
\triangleq
\mathbb{E}
\left[
\hat{\mathbf{z}}_i^{(b)}
\mid
\lambda_i,\mathcal{O}_i
\right]
\approx
\lambda_i
\tilde{\mathbf{z}}_i^{(b)}
+
\left(1-\lambda_i\right)
\mathbf{s}_i^{(a\rightarrow b)} ,
\label{eq:posterior_mean}
\end{equation}
where $\tilde{\mathbf{z}}_i^{(b)}$ denotes the counterpart representation in the target view, 
$\mathbf{s}_i^{(a\rightarrow b)}
=
\sum_{c=1}^{C}
q_{ic}^{(a\rightarrow b)}
\boldsymbol{\mu}_{c}^{(b)}$
denotes the transported semantic counterpart obtained from prototype-level semantic transport, and $\lambda_i$ controls their relative contributions.
Under the imperfect information problem, the counterpart may be unreliable or unavailable, resulting in the following two situations:
% \begin{equation}
% \left(\tilde{\mathbf{z}}_i^{(b)}, \lambda_i\right)
% =
% \left\{
% \begin{array}{ll}
% \left(\mathbf{z}_i^{(b)},\ \gamma_i
% \left\langle
% \mathbf{q}_i^{(a\rightarrow b)},
% \mathbf{q}_i^{(b)}
% \right\rangle\right),
% & m_i^{(b)}=1,\\[2mm]
% \left(g^{(a\rightarrow b)}(\mathbf{z}_i^{(a)}),\ 1/2\right),
% & m_i^{(b)}=0,
% \end{array}
% \right.
% \label{eq:imperfect_counterpart_lambda}
% \end{equation}
% where $\gamma_i$ is the estimated correspondence reliability, 
% $\left\langle
% \mathbf{q}_i^{(a\rightarrow b)},
% \mathbf{q}_i^{(b)}
% \right\rangle$
% measures the semantic consistency between the transported prototype assignment and the observed target-view assignment, and
% $g^{(a\rightarrow b)}(\cdot)$ is the cross-view generator from view $a$ to view $b$, following the common practice of IV-oriented MvC methods~\cite{divide,completer}.

\begin{itemize}
  \item \textbf{For complete but reliability-agnostic instances.} The target-view observation is available, and $\tilde{\mathbf{z}}_i^{(b)}$ is given by the observed target-view representation $\mathbf{z}_i^{(b)}$. However, this observed counterpart may be mismatched with the anchor view due to noisy correspondences. 
  To alleviate this issue, $\lambda_i=\gamma_i
  \left\langle
  \mathbf{q}_i^{(a\rightarrow b)},
  \mathbf{q}_i^{(b)}
  \right\rangle$ serves as an adaptive weight in Eq.~\ref{eq:posterior_mean}, controlling the relative contribution of the observed counterpart and the transported semantic counterpart. Specifically, it combines the estimated correspondence reliability $\gamma_i$ with the semantic consistency $\left\langle \mathbf{q}_i^{(a\rightarrow b)}, \mathbf{q}_i^{(b)} \right\rangle$ between the transported prototype assignment and the observed target-view assignment. 
  Consequently, reliable complete instances assign larger weights to the observed counterpart, whereas unreliable ones are calibrated by the transported semantic counterpart.
  \item \textbf{For incomplete and availability-deficient instances.}
  The target-view observation is unavailable, and thus $\mathbf{z}_i^{(b)}$ cannot be directly observed. Accordingly, $\tilde{\mathbf{z}}_i^{(b)}$ has to be constructed from the anchor-view representation through the cross-view generator $g^{(a\rightarrow b)}(\mathbf{z}_i^{(a)})$, following the common practice of IV-oriented MvC methods~\cite{divide,completer}. However, this generated counterpart may be affected by inaccurate cross-view mappings, especially when complete instances are contaminated by noisy correspondences, as depicted in Fig.~\ref{fig1}. To reduce the dependence on the generated counterpart alone, $\lambda_i$ is set to $1/2$ in Eq.~\ref{eq:posterior_mean}, allowing the generated counterpart and the transported semantic counterpart to contribute equally to the inferred latent counterpart. In this way, the inferred latent counterpart is constrained by the target-view semantic structure rather than relying solely on the generated counterpart.
\end{itemize}

With the inferred counterparts, one could optimize PLCI by the following joint objective:
\begin{equation}
\mathcal{L}_{\mathrm{PLCI}}
=
\mathcal{L}_{\mathrm{NC}}
+
\alpha
\mathcal{L}_{\mathrm{IV}},
\label{eq:l_plci}
\end{equation}
where $\alpha$ is a trade-off hyperparameter fixed as $0.2$ in our experiments. 
$\mathcal{L}_{\mathrm{NC}}$ and $\mathcal{L}_{\mathrm{IV}}$ are designed for learning with noisy correspondences and incomplete views, respectively.

Specifically, $\mathcal{L}_{\mathrm{NC}}$ is imposed on complete instances and thus given by:
\begin{equation}
\mathcal{L}_{\mathrm{NC}}
=
\sum
\ell_{\mathrm{con}}
\left(f^{(a)}(\mathbf{x}_i^{(a)}), \hat{\mathbf{z}}_i^{(b)}  \right),
\label{eq:l_nc}
\end{equation}
where $f^{(a)}(\cdot)$ is the view-specific encoder, $\ell_{\mathrm{con}}(\cdot, \cdot)$ denotes the cross-view contrastive loss~\cite{oord2018representation}. 
By utilizing the inferred latent counterpart, $\mathcal{L}_{\mathrm{NC}}$ encourages robust cross-view alignment under noisy correspondences.

As for incomplete instances, instead of directly applying to the inferred counterparts, the proposed method manipulates the transported semantic structure to regularize the inferred counterparts.
Formally,
\begin{equation}
\mathcal{L}_{\mathrm{IV}}
=
\sum
D_{\mathrm{KL}}
\left(
\mathbf{q}_i^{(a\rightarrow b)}
\|
\tilde{\mathbf{q}}_i^{(b)}
\right),
\label{eq:l_iv}
\end{equation}
where $D_{\mathrm{KL}}(\cdot\|\cdot)$ denotes the KL divergence, $\mathbf{q}_i^{(a\rightarrow b)}$ is the transported prototype assignment defined in Eq.~\ref{eq:transported_assignment}, and $\tilde{\mathbf{q}}_i^{(b)}$ is the prototype assignment of the inferred latent counterpart $\hat{\mathbf{z}}_i^{(b)}$. 
By minimizing $\mathcal{L}_{\mathrm{IV}}$, the inferred counterpart is encouraged to preserve the transported target-view semantic structure, thereby facilitating robust clustering when target-view observations are missing.

\section{Experiments}
\label{Experiments}
In this section, we carry out extensive experiments to evaluate the effectiveness of the proposed PLCI in addressing the imperfect information problem.
Due to space limitations, we present more details and results in the supplementary material.

\subsection{Experimental Setup}
\paragraph{Datasets.} 
We evaluate our method on six widely used multi-view benchmarks including Scene15~\cite{scene15}, LandUse21~\cite{landuse21}, Reuters~\cite{reuters}, CCV20~\cite{ccv20}, HandWritten~\cite{handwritten}, and SUN RGB-D~\cite{bml}. 
These datasets cover a wide range of modalities, including video-audio, RGB-depth, image-text, and multilingual text, thus providing a comprehensive testbed for evaluating the robustness of MvC methods against imperfect information.

\begin{table*}[htbp]
\caption{Clustering performance comparisons on five widely-used multi-view datasets under different IIR. The results are the mean of five runs. The best and second-best results are marked as \textbf{bold} and \underline{underline}, respectively. The background colors \colorbox{cyan!10}{\phantom{xx}}, \colorbox{orange!10}{\phantom{xx}}, and \colorbox{violet!10}{\phantom{xx}} indicate methods designed for handling incomplete information, incomplete views, and noisy correspondences, respectively.}
\label{tab:main_results}
\centering
\scriptsize
\setlength{\tabcolsep}{2.5pt}
\renewcommand{\arraystretch}{1.1}
\resizebox{\textwidth}{!}{%
\begin{tabular}{@{}ll*{18}{c}@{}}
\toprule
\multirow{2}{*}{Setting} & \multicolumn{1}{c}{\multirow{2}{*}{Method}}
& \multicolumn{3}{c}{Scene15}
& \multicolumn{3}{c}{LandUse21}
& \multicolumn{3}{c}{Reuters}
& \multicolumn{3}{c}{CCV20}
& \multicolumn{3}{c}{HandWritten}
& \multicolumn{3}{c}{Average} \\
\cmidrule(lr){3-5}
\cmidrule(lr){6-8}
\cmidrule(lr){9-11}
\cmidrule(lr){12-14}
\cmidrule(lr){15-17}
\cmidrule(lr){18-20}
& & ACC & NMI & ARI & ACC & NMI & ARI & ACC & NMI & ARI & ACC & NMI & ARI & ACC & NMI & ARI & ACC & NMI & ARI \\
\midrule
\multirow{9}{*}{\shortstack{IIR=0}} 
 & \cellcolor{cyan!10}SURE (TPAMI'23) & 40.60 & 42.46 & 24.23 & 26.59 & 30.78 & 12.90 & 49.12 & 30.74 & 24.70 & 19.94 & 16.82 & 6.31 & 66.73 & 60.15 & 49.01 & 40.60 & 36.19 & 23.43 \\
 & \cellcolor{cyan!10}CAMERA (TPAMI'26) & 45.98 & 46.66 & 28.88 & 29.32 & 35.97 & 15.73 & 52.80 & 32.16 & 26.48 & \second{23.30} & \second{19.47} & \second{8.09} & 92.17 & 87.96 & 85.35 & 48.71 & 44.44 & 32.91 \\
 & \cellcolor{orange!10}GCFAgg (CVPR'23) & 42.34 & 41.96 & 24.46 & 26.59 & 30.31 & 12.79 & 32.19 & 15.35 & 7.79 & 19.16 & 15.89 & 5.75 & 92.41 & 86.10 & 84.14 & 42.54 & 37.92 & 26.99 \\
 & \cellcolor{orange!10}DIVIDE (AAAI'24) & 45.17 & 45.74 & 28.62 & \second{32.17} & \best{39.71} & 17.92 & \second{53.40} & \best{36.64} & 27.96 & 22.89 & 17.51 & 7.24 & \best{94.48} & \best{88.55} & \best{88.17} & \second{49.62} & \second{45.63} & \second{33.98} \\
 & \cellcolor{orange!10}FreeCSL (CVPR'25) & 41.54 & 44.18 & 25.02 & 28.51 & 34.03 & 15.29 & 48.92 & 32.59 & \second{28.14} & 21.39 & 17.60 & 6.86 & 81.77 & 83.83 & 76.78 & 44.43 & 42.45 & 30.42 \\
 & \cellcolor{violet!10}CANDY (NeurIPS'24) & 41.70 & 41.10 & 24.84 & 30.63 & 35.95 & 16.11 & 48.43 & 32.57 & 24.67 & 20.88 & 16.54 & 6.85 & 90.21 & 82.28 & 79.88 & 46.37 & 41.69 & 30.47 \\
 & \cellcolor{violet!10}ROLL (CVPR'25) & \best{47.10} & \best{48.96} & \best{30.13} & 29.15 & 34.55 & 15.16 & 47.06 & 23.12 & 22.47 & 23.10 & 19.09 & 7.84 & 73.85 & 56.12 & 51.93 & 44.05 & 36.37 & 25.51 \\
 & \cellcolor{violet!10}CorreGen (ICLR'26) & 45.62 & 45.83 & 29.09 & 31.74 & \second{38.72} & \second{18.05} & 46.41 & 29.04 & 20.74 & 22.45 & 19.26 & 7.95 & 92.33 & 86.14 & 84.05 & 47.71 & 43.80 & 31.98 \\
 & PLCI (Ours) & \second{46.55} & \second{46.73} & \second{29.86} & \best{32.35} & 38.40 & \best{18.52} & \best{56.37} & \second{35.12} & \best{29.22} & \best{25.25} & \best{20.85} & \best{9.09} & \second{93.97} & \second{88.40} & \second{87.12} & \best{50.90} & \best{45.90} & \best{34.76} \\
\midrule
\multirow{11}{*}{\shortstack{IIR=0.2}} 
 & \cellcolor{cyan!10}SURE (TPAMI'23) & 38.95 & 36.07 & 20.45 & 24.31 & 25.61 & 10.02 & 40.66 & 17.44 & 12.97 & 18.39 & 13.43 & 4.93 & 67.15 & 57.83 & 48.49 & 37.89 & 30.08 & 19.37 \\
 & \cellcolor{cyan!10}SURE* (TPAMI'23) & 39.11 & 40.82 & 23.18 & 26.82 & 30.49 & 13.30 & 44.69 & 27.49 & 21.53 & 20.07 & 16.73 & 6.39 & 73.22 & 64.91 & 57.82 & 40.78 & 36.09 & 24.45 \\
 & \cellcolor{cyan!10}CAMERA (TPAMI'26) & 41.83 & 40.28 & 23.72 & 25.95 & 29.14 & 11.71 & 51.82 & 28.58 & \second{25.22} & \second{21.28} & 16.46 & 6.64 & 77.37 & 64.76 & 61.67 & 43.65 & 35.84 & 25.79 \\
 & \cellcolor{cyan!10}CAMERA* (TPAMI'26) & 42.57 & 40.99 & 24.23 & 25.63 & 28.40 & 11.45 & 50.75 & 28.63 & 25.1 & 21.22 & 16.52 & 6.69 & 79.72 & 64.61 & 61.98 & 43.98 & 35.83 & 25.89 \\
 & \cellcolor{orange!10}GCFAgg (CVPR'23) & 38.02 & 34.76 & 19.97 & 25.00 & 26.93 & 11.44 & 29.79 & 10.98 & 4.76 & 17.76 & 13.39 & 4.82 & 79.78 & 70.81 & 66.80 & 38.07 & 31.37 & 21.56 \\
 & \cellcolor{orange!10}DIVIDE (AAAI'24) & 43.01 & 40.82 & 25.90 & 29.94 & 34.75 & 15.84 & 51.51 & 30.21 & 25.31 & 20.59 & 13.85 & 5.60 & 88.00 & 75.70 & 75.36 & \second{46.61} & 39.07 & \second{29.60} \\
 & \cellcolor{orange!10}FreeCSL (CVPR'25) & 38.18 & 37.43 & 22.07 & 27.94 & 30.58 & 13.74 & 44.10 & 23.39 & 20.74 & 19.69 & 14.86 & 5.83 & 73.61 & 66.65 & 61.60 & 40.70 & 34.58 & 24.80 \\
 & \cellcolor{violet!10}CANDY (NeurIPS'24) & 40.11 & 39.68 & 24.04 & 27.74 & 29.09 & 13.44 & \second{52.47} & \best{33.59} & 26.77 & 18.52 & 14.33 & 5.57 & 88.13 & 78.28 & 75.98 & 45.39 & 38.99 & 29.16 \\
 & \cellcolor{violet!10}ROLL (CVPR'25) & \second{45.52} & \second{44.47} & \second{27.51} & 28.75 & 32.04 & 14.25 & 45.94 & 20.61 & 20.00 & 20.86 & 15.63 & 6.23 & 79.90 & 64.14 & 61.12 & 44.19 & 35.38 & 25.82 \\
 & \cellcolor{violet!10}CorreGen (ICLR'26) & 41.63 & 43.05 & 26.08 & \second{30.25} & \second{35.92} & \second{16.70} & 38.53 & 21.51 & 13.24 & 21.20 & \second{17.52} & \second{6.93} & \second{89.56} & \second{80.65} & \second{78.66} & 44.23 & \second{39.73} & 28.32 \\
 & PLCI (Ours) & \best{46.17} & \best{45.20} & \best{28.62} & \best{32.04} & \best{36.27} & \best{17.58} & \best{54.58} & \second{32.22} & \best{28.51} & \best{24.29} & \best{19.94} & \best{8.54} & \best{90.97} & \best{81.84} & \best{81.24} & \best{49.61} & \best{43.09} & \best{32.90} \\
\midrule
\multirow{11}{*}{\shortstack{IIR=0.5}} 
 & \cellcolor{cyan!10}SURE (TPAMI'23) & 32.75 & 31.13 & 16.53 & 19.20 & 18.65 & 5.98 & 39.64 & 18.14 & 14.80 & 13.64 & 9.23 & 2.86 & 53.28 & 41.46 & 32.05 & 31.70 & 23.72 & 14.44 \\
 & \cellcolor{cyan!10}SURE* (TPAMI'23) & 38.93 & 39.28 & 22.97 & 23.58 & 26.32 & 10.52 & 45.71 & 27.61 & 22.09 & 18.27 & 15.17 & 5.52 & 66.91 & 58.45 & 49.94 & 38.68 & 33.37 & 22.21 \\
 & \cellcolor{cyan!10}CAMERA (TPAMI'26) & 33.57 & 33.02 & 17.30 & 20.66 & 21.10 & 7.10 & 45.63 & 21.39 & 19.01 & 17.28 & 12.50 & 4.70 & 68.63 & 48.16 & 44.27 & 37.15 & 27.23 & 18.48 \\
 & \cellcolor{cyan!10}CAMERA* (TPAMI'26) & 36.96 & 36.04 & 19.46 & 21.42 & 22.71 & 7.84 & 48.35 & 24.68 & 20.85 & 19.04 & 13.74 & 5.41 & 71.12 & 51.79 & 47.91 & 39.38 & 29.79 & 20.29 \\
 & \cellcolor{orange!10}GCFAgg (CVPR'23) & 32.82 & 30.17 & 16.35 & 21.72 & 21.74 & 8.14 & 32.55 & 10.65 & 6.88 & 16.21 & 11.04 & 3.79 & 58.55 & 45.34 & 37.83 & 32.37 & 23.79 & 14.60 \\
 & \cellcolor{orange!10}DIVIDE (AAAI'24) & 37.51 & 32.52 & 19.83 & 27.14 & 28.29 & 12.68 & \best{51.45} & \second{26.37} & \second{22.72} & 15.25 & 8.98 & 2.89 & 77.89 & 61.58 & 59.23 & 41.85 & 31.55 & 23.47 \\
 & \cellcolor{orange!10}FreeCSL (CVPR'25) & 32.03 & 28.24 & 15.76 & 22.73 & 22.11 & 8.83 & 34.42 & 10.45 & 9.31 & 15.54 & 10.18 & 3.33 & 66.99 & 52.88 & 48.16 & 34.34 & 24.77 & 17.08 \\
 & \cellcolor{violet!10}CANDY (NeurIPS'24) & 36.40 & 36.31 & 20.94 & 22.46 & 22.44 & 8.85 & 42.46 & 22.23 & 16.53 & 15.23 & 10.38 & 3.49 & 79.67 & 64.75 & 61.02 & 39.24 & 31.22 & 22.17 \\
 & \cellcolor{violet!10}ROLL (CVPR'25) & \second{39.85} & 36.68 & \second{21.89} & 25.45 & 27.25 & 11.60 & 40.39 & 15.34 & 14.90 & 16.61 & 10.43 & 3.82 & 72.30 & 57.21 & 50.82 & 38.92 & 29.38 & 20.61 \\
 & \cellcolor{violet!10}CorreGen (ICLR'26) & 38.31 & \second{38.13} & \second{21.89} & \second{27.27} & \second{30.02} & \second{12.80} & {45.04} & 24.79 & 18.83 & \second{19.11} & \second{14.77} & \second{5.51} & \second{80.50} & \second{68.49} & \second{64.36} & \second{42.05} & \second{35.24} & \second{24.68} \\
 & PLCI (Ours) & \best{42.17} & \best{41.39} & \best{25.00} & \best{30.09} & \best{34.58} & \best{16.13} & \second{49.80} & \best{26.93} & \best{23.56} & \best{22.97} & \best{18.78} & \best{8.06} & \best{86.94} & \best{76.08} & \best{74.05} & \best{46.39} & \best{39.55} & \best{29.36} \\
\midrule
\multirow{11}{*}{\shortstack{IIR=0.8}} 
 & \cellcolor{cyan!10}SURE (TPAMI'23) & 21.81 & 18.34 & 8.07 & 14.72 & 12.53 & 3.23 & 31.41 & 10.49 & 8.17 & 11.83 & 6.26 & 1.8 & 37.54 & 26.93 & 17.43 & 23.47 & 14.91 & 7.74 \\
 & \cellcolor{cyan!10}SURE* (TPAMI'23) & 23.92 & 20.60 & 9.30 & 15.46 & 13.87 & 3.65 & \second{43.02} & \best{23.25} & \second{18.16} & 14.17 & 9.81 & 3.15 & 41.63 & 34.42 & 22.05 & 27.64 & 20.39 & 11.26 \\
 & \cellcolor{cyan!10}CAMERA (TPAMI'26) & 27.44 & 28.06 & 12.94 & 18.27 & 18.95 & 5.37 & 38.13 & 13.54 & 12.03 & 13.70 & 8.45 & 2.61 & 46.99 & 34.25 & 24.28 & \second{28.91} & 20.65 & 11.45 \\
 & \cellcolor{cyan!10}CAMERA* (TPAMI'26) & 22.94 & 23.42 & 9.08 & 15.04 & 18.93 & 4.26 & \best{45.15} & \second{22.06} & \best{19.06} & 15.93 & 11.02 & 3.82 & 30.17 & 27.19 & 12.35 & 25.85 & 20.52 & 9.71 \\
 & \cellcolor{orange!10}GCFAgg (CVPR'23) & 26.49 & 23.77 & 11.39 & 16.36 & 15.75 & 4.48 & 28.26 & 6.71 & 4.45 & 12.54 & 8.07 & 2.31 & 42.81 & 36.82 & 23.52 & 25.29 & 18.22 & 9.23 \\
 & \cellcolor{orange!10}DIVIDE (AAAI'24) & \second{29.24} & 21.83 & 11.93 & 20.29 & 20.73 & 7.14 & 28.92 & 8.18 & 6.11 & 11.28 & 5.16 & 1.35 & \second{51.61} & 36.64 & \second{28.27} & 28.27 & 18.51 & 10.96 \\
 & \cellcolor{orange!10}FreeCSL (CVPR'25) & 23.04 & 18.31 & 8.62 & 17.73 & 15.23 & 4.79 & 29.46 & 10.02 & 6.40 & 13.48 & 8.23 & 2.22 & 38.86 & 22.22 & 15.47 & 24.51 & 14.80 & 7.50 \\
 & \cellcolor{violet!10}CANDY (NeurIPS'24) & 28.46 & 28.36 & 13.82 & 17.46 & 18.11 & 5.64 & 31.89 & 11.86 & 7.57 & 13.60 & 8.57 & 2.51 & 52.98 & 37.55 & 29.48 & 28.88 & 20.89 & 11.80 \\
 & \cellcolor{violet!10}ROLL (CVPR'25) & 28.65 & 24.45 & 12.68 & 19.38 & 16.78 & 6.35 & 34.33 & 10.71 & 9.94 & 11.14 & 4.43 & 1.14 & 46.21 & 34.95 & 21.03 & 27.94 & 18.26 & 10.23 \\
 & \cellcolor{violet!10}CorreGen (ICLR'26) & 28.12 & \second{31.41} & \second{14.99} & \second{21.15} & \second{24.67} & \second{8.58} & 34.73 & 15.44 & 10.53 & 12.46 & \second{10.89} & \second{3.05} & 43.32 & \second{39.60} & 25.70 & 27.96 & \second{24.40} & \second{12.57} \\
 & PLCI (Ours) & \best{37.78} & \best{36.71} & \best{22.21} & \best{26.66} & \best{30.68} & \best{11.68} & 40.72 & 18.56 & 15.00 & \best{16.47} & \best{12.39} & \best{4.35} & \best{72.54} & \best{63.07} & \best{57.13} & \best{38.83} & \best{32.28} & \best{22.07} \\
\bottomrule
\end{tabular}%
}
\end{table*}

\paragraph{Implementation Details.}
PLCI is a general framework that can enhance existing MvC methods with robustness against imperfect information. 
In this work, we verify the effectiveness of PLCI on top of the recently proposed DIVIDE~\cite{divide}. 
Specifically, to endow the backbone with robustness, we retain the backbone and training pipeline of DIVIDE, and incorporate the proposed instance-level reliability estimation, prototype-level semantic transport, and latent-counterpart-based joint objective.
To evaluate robustness against imperfect information, we simulate different imperfect information rate (IIR) settings by randomly introducing incomplete views and noisy correspondences into the original datasets. 
Specifically, for each instance, we randomly mask its views with a certain probability to simulate incomplete views, and randomly shuffle the correspondences of a certain proportion of complete instances to simulate noisy correspondences.

\subsection{Comparison with State of the Arts}
We compare our method with 10 state-of-the-art MvC methods, including the IV-oriented methods (GCFAgg~\cite{gcfagg}, DIVIDE~\cite{divide}, and FreeCSL~\cite{freecsl}), the NC-oriented methods (CANDY~\cite{candy}, ROLL~\cite{roll}, and CorreGen~\cite{corregen}), and the methods designed for handling incomplete information (SURE~\cite{sure} and CAMERA~\cite{camera}).
In addition, we report the results of SURE and CAMERA using instances with fully clean correspondences, which serve as quite strong baselines, as the used data does not contain any noise.
Following widely used evaluation protocols, we adopt clustering accuracy (ACC), normalized mutual information (NMI), and adjusted Rand index (ARI) as evaluation metrics.

Table~\ref{tab:main_results} summarizes the clustering performance under four IIR settings, where one could observe that the proposed PLCI consistently achieves the best average performance across all settings, demonstrating its superior robustness against imperfect information.

\subsection{Experiments on Raw Data Inputs}

To further assess whether PLCI can be applied beyond pre-extracted multi-view features, we conduct additional experiments on SUN RGB-D with raw visual inputs. The two views are RGB images and depth maps. We replace the original feature encoder with a ResNet-18 backbone for each view, while keeping the remaining structure of PLCI unchanged. We compare PLCI with DIVIDE and CorreGen, which are the most competitive IV- and NC-oriented baselines in the main experiments.

As shown in Table~\ref{tab:sunrgbd_raw}, PLCI remains competitive under the clean setting and achieves the best NMI and ARI. When NC and IV coexist, PLCI consistently outperforms both baselines across all metrics. The advantage becomes more pronounced as the degradation increases. Under the most challenging setting, \ie, IIR=0.8, PLCI improves over the second-best method by 4.19, 3.99, and 2.07 in ACC, NMI, and ARI, respectively. These results indicate that PLCI is not restricted to pre-extracted feature inputs and can also work effectively with raw data inputs.

\begin{table}[htbp]
\centering
\scriptsize
\setlength{\tabcolsep}{3.0pt}
\renewcommand{\arraystretch}{1.1}
\caption{Clustering performance comparisons on SUN RGB-D dataset under different NC and IV settings. The best and second-best results are shown in \textbf{bold} and \underline{underline}, respectively.}
\label{tab:sunrgbd_raw}
\resizebox{0.9\textwidth}{!}{%
\begin{tabular}{@{}ll cc>{\columncolor{gray!20}}c cc>{\columncolor{gray!20}}c cc>{\columncolor{gray!20}}c cc>{\columncolor{gray!20}}c@{}}
\toprule
\multirow{2}{*}{Dataset} & \multirow{2}{*}{Metric}
& \multicolumn{3}{c}{IIR=0}
& \multicolumn{3}{c}{IIR=0.2}
& \multicolumn{3}{c}{IIR=0.5}
& \multicolumn{3}{c}{IIR=0.8} \\
\cmidrule(lr){3-5}
\cmidrule(lr){6-8}
\cmidrule(lr){9-11}
\cmidrule(lr){12-14}
& & DIVIDE & CorreGen & Ours
  & DIVIDE & CorreGen & Ours
  & DIVIDE & CorreGen & Ours
  & DIVIDE & CorreGen & Ours \\
\midrule
\multirow{3}{*}{SUN RGB-D}
& ACC
& \textbf{35.95} & 34.90 & \underline{34.98}
& 32.00 & \underline{33.46} & \textbf{34.06}
& \underline{26.80} & 25.13 & \textbf{28.27}
& 22.21 & \underline{23.93} & \textbf{28.12} \\
& NMI
& 34.30 & \underline{34.53} & \textbf{35.04}
& 27.40 & \underline{28.77} & \textbf{31.93}
& \underline{19.91} & 19.81 & \textbf{22.68}
& 18.21 & \underline{20.56} & \textbf{24.55} \\
& ARI
& \underline{22.14} & 20.70 & \textbf{22.21}
& 17.15 & \underline{17.93} & \textbf{20.17}
& \underline{12.51} & 10.65 & \textbf{13.08}
& 8.92 & \underline{10.39} & \textbf{12.46} \\
\bottomrule
\end{tabular}%
}
\end{table}

\subsection{Ablation Studies}

To investigate the effectiveness of each component in PLCI, we conduct ablation studies on three datasets, \ie, Scene15, LandUse21, and CCV20, under two IIR settings, \ie, IIR$=0.5$ and IIR$=0.8$.
More specifically, we compare the following variants of PLCI: i) w/o $\mathcal{L}_{\mathrm{NC}}$: PLCI without the prototype-guided counterpart correction for unreliable correspondences, which directly applies the cross-view contrastive loss on the observed counterparts; ii) w/o $\mathcal{L}_{\mathrm{IV}}$: PLCI without the prototype-guided counterpart substitution for missing views, which does not utilize the inferred latent counterparts for regularization; iii) Full PLCI: the complete version of PLCI with all components used.
The results are shown in Table~\ref{tab:ablation}, where one could observe that each component serves as an inseparable part of PLCI for achieving robust MvC against imperfect information.

\subsection{Study on the Generalizability}
\begin{figure}[htbp]
  \centering
  
  % ================= 左侧：表格 =================
  \begin{minipage}[c]{0.62\textwidth}
    \centering
    % 使用 captionof 生成 Table 标题
    \captionof{table}{Ablation studies of PLCI on datasets Scene15, LandUse21 and CCV20. \cmark\ denotes that the corresponding module is used.}
    \label{tab:ablation}
    \scriptsize
    \setlength{\tabcolsep}{3.2pt}
    \renewcommand{\arraystretch}{1.1}
    % 注意这里把 0.75\columnwidth 改为了 \linewidth，让表格自适应 minipage 的宽度
    \resizebox{\linewidth}{!}{%
    \begin{tabular}{@{}lcc*{9}{c}@{}}
    \toprule
    \multirow{2}{*}{Setting} 
    & \multirow{2}{*}{$\mathcal{L}_{\mathrm{NC}}$} 
    & \multirow{2}{*}{$\mathcal{L}_{\mathrm{IV}}$}
    & \multicolumn{3}{c}{Scene15}
    & \multicolumn{3}{c}{LandUse21}
    & \multicolumn{3}{c}{CCV20} \\
    \cmidrule(lr){4-6}
    \cmidrule(lr){7-9}
    \cmidrule(lr){10-12}
    & & & ACC & NMI & ARI & ACC & NMI & ARI & ACC & NMI & ARI \\
    \midrule
    \multirow{4}{*}{\shortstack{IIR=0.5}}
    % &        &        & 30.43 & 33.36 & 17.21 & 17.70 & 18.52 & 5.55  & 17.59 & 13.77 & 5.35 \\
    & \cmark &        & \second{32.99} & \second{34.71} & \second{17.99} & \second{24.17} & \second{28.11} & \second{10.96} & \second{19.11} & \second{15.27} & \second{6.00} \\
    &        & \cmark & 31.57 & 30.36 & 15.66 & 23.75 & 26.32 & 10.17 & 15.47 & 10.20 & 3.53 \\
    & \cmark & \cmark & \best{42.17} & \best{41.39} & \best{25.00} & \best{30.09} & \best{34.58} & \best{16.13} & \best{22.97} & \best{18.78} & \best{8.06} \\
    \midrule
    \multirow{4}{*}{\shortstack{IIR=0.8}}
    % &        &        & 23.42 & 23.22 & 10.03 & 17.24 & 17.79 & 5.27  & 11.51 & 5.79  & 1.60 \\
    & \cmark &        & \second{28.80} & \second{31.24} & \second{15.09} & \second{20.38} & \second{23.18} & 7.71 & \second{14.31} & \second{9.76} & \second{3.12} \\
    &        & \cmark & 24.40 & 21.90 & 9.97 & 20.34 & 23.12 & \second{7.80} & 12.45 & 6.69 & 1.93 \\
    & \cmark & \cmark & \best{37.78} & \best{36.71} & \best{22.21} & \best{26.66} & \best{30.68} & \best{11.68} & \best{16.47} & \best{12.39} & \best{4.35} \\
    \bottomrule
    \end{tabular}%
    }
  \end{minipage}\hfill % \hfill 用于推开左右两边的 minipage，留出中间的间隙
  % ================= 右侧：图片 =================
  \begin{minipage}[c]{0.35\textwidth}
    \centering
    % 注意这里把 width 改成了 \linewidth，使得图片填满这个 minipage
    \includegraphics[width=\linewidth]{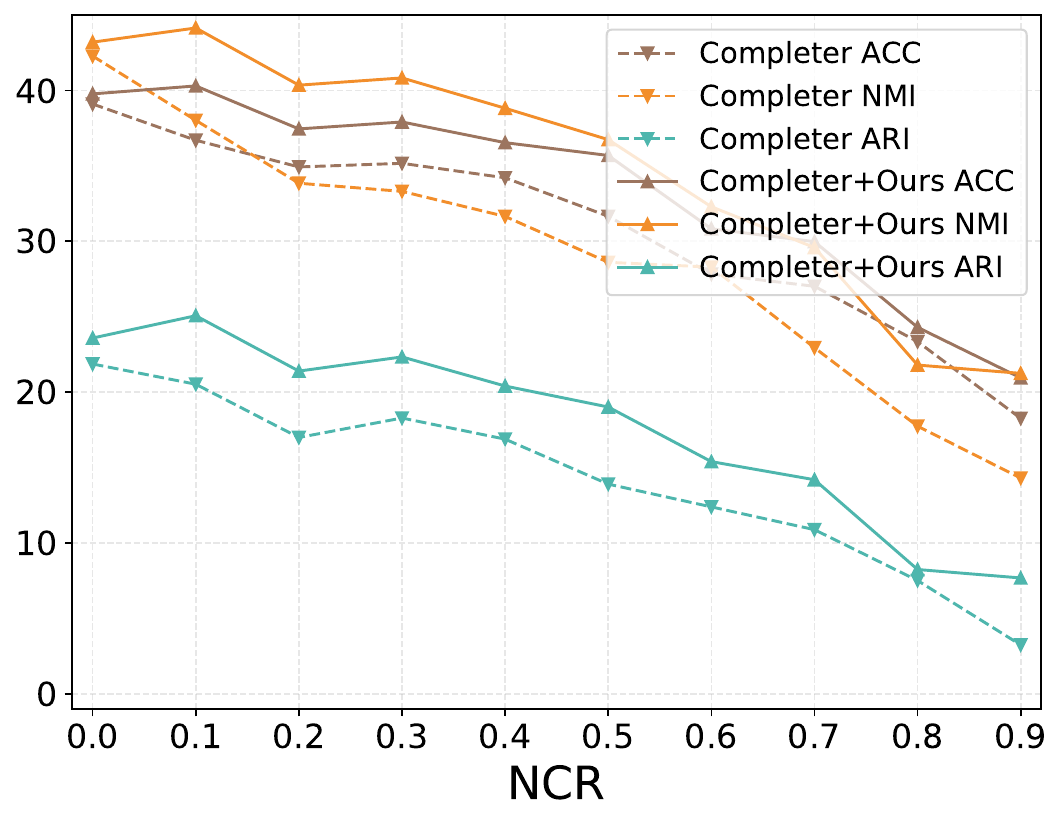} 
    \caption{Generalizability study on Scene15.}
    \label{fig:generalizability}
  \end{minipage}
  
\end{figure}

To verify the generalizability of the proposed method, we further apply PLCI to another representative MvC method, Completer~\cite{completer}.
For a comprehensive evaluation, we fix the incomplete view rate (IVR) at 0.5 and vary the noisy correspondence rate (NCR) from 0.0 to 0.9 with an interval of 0.1 on Scene15.
As shown in Fig.~\ref{fig:generalizability}, Completer+Ours consistently outperforms Completer across all NCR settings, demonstrating the effectiveness of PLCI in enhancing the robustness of existing MvC methods against imperfect information.

\section{Conclusion}
In this paper, we study the imperfect information problem in multi-view clustering (MvC), where incomplete views (IV) and noisy correspondences (NC) arise simultaneously in real-world multi-view data. 
We demonstrate that these two challenges could be unified from the perspective of imperfect cross-view counterpart information, in which the desired counterpart of an anchor instance may be either unavailable or unreliable. 
Based on this observation, we propose Posterior-guided Latent Counterpart Inference (PLCI), a robust MvC framework that formulates the desired counterpart as a latent variable and infers it by jointly exploiting instance-level reliability and prototype-level semantic transport. Extensive experiments on widely used multi-view benchmarks demonstrate that PLCI consistently outperforms state-of-the-art methods under diverse imperfect information settings. 
As the multi-modal/view data has an incentive to be contaminated by imperfect information, in the future, we plan to extend the imperfect information problem to more complex scenarios and applications, including but not limited to cross-modal retrieval, multi-modal person re-identification, and multi-view representation learning.

\medskip
{
\bibliographystyle{plainnat}
\bibliography{neurips_2026}
}

%%%%%%%%%%%%%%%%%%%%%%%%%%%%%%%%%%%%%%%%%%%%%%%%%%%%%%%%%%%%

\appendix
\section{Related Work}
In this section, we briefly review two most related topics to this work, \ie, multi-view clustering with incomplete views and multi-view clustering with noisy correspondence. 

\subsection{Multi-View Clustering with Incomplete Views}

According to whether the missing views are explicitly recovered, existing IV-oriented MvC methods~\cite{qin2023nim,daimc,apadc,completer,divide,fu2022unified,chen2020multi,chen2022efficient} can be roughly categorized into two groups, \ie, recovery-free and recovery-based methods. 
In brief, recovery-free methods aim to learn a common representation from the available views and then perform clustering in the shared representation space.
For example, DAIMC~\cite{daimc} aligns view-specific matrices and latent representations via weighted non-negative matrix factorization to learn a consensus representation from incomplete data. 
APADC~\cite{apadc} performs adaptive feature projection and distribution alignment to reduce the cross-view discrepancy among available observations. 
Different from them, recovery-based methods explicitly infer the unavailable views.
COMPLETER~\cite{completer} recovers missing representations through cross-view contrastive prediction, 
while DCP~\cite{dcp} further unifies consistency learning and missing-view recovery into a dual contrastive prediction framework. 
DIVIDE~\cite{divide} employs cross-view decoders to recover missing samples and performs decoupled contrastive learning to preserve both cross-view consistency and view-specific information. 
Although these methods have achieved promising performance under incomplete views, they generally rely on reliable cross-view correspondences to learn cross-view recovery. 
Once the observed correspondences are noisy, the learned recovery could be seriously deviated from the desired one, thus leading to erroneous recovery. 

Different from existing IV-oriented methods that mainly handle missing views under reliable cross-view correspondences, this study addresses a more practical problem where incomplete views and noisy correspondences coexist. Rather than reconstructing missing views or explicitly re-aligning correspondences, our method models the desired counterpart of each instance as a latent variable and infers its posterior distribution by jointly exploiting instance-level reliability and prototype-level semantic transport. This enables unavailable and unreliable counterparts to be handled in a unified probabilistic framework.

\subsection{Multi-View Clustering with Noisy Correspondence}
Besides instance completeness, another key assumption in conventional MvC is view consistency, which treats cross-view counterparts as correctly aligned. 
Different from partially view-aligned problem (PVP), where a subset of aligned instances is known during training and can serve as reliable anchors~\cite{sure,camera}, NC-oriented MvC assumes that the reliability of observed counterparts is unknown. Existing methods mainly improve clustering by suppressing the adverse effect of unreliable pairs during representation learning~\cite{roll,rmcnc}, handling false positive and false negative correspondences in contrastive supervision~\cite{candy}, or estimating soft correspondence distributions under a generative framework~\cite{corregen}.

However, these methods mainly focus on unreliable observed counterparts while assuming that all views are available. As a result, instances with unavailable views are usually excluded from their formulation, and the information from unavailable views cannot be exploited during training. Therefore, existing NC-oriented MvC methods mainly target unreliable counterparts, but are not designed for the setting considered in this work, where unavailable views and unreliable counterparts coexist.

\section{More Experiment Details}
\subsection{Details of Used Datasets}
\begin{itemize}
    \item \textbf{Scene15}~\cite{scene15} contains 4,485 images from 15 indoor and outdoor scene categories. Following~\cite{sure}, we use the PHOG and GIST features as two views.

    \item \textbf{LandUse21}~\cite{landuse21} consists of 2,100 satellite images from 21 categories. Following~\cite{dcp}, we use the PHOG and LBP features as two views.

    \item \textbf{Reuters}~\cite{reuters} is a multilingual news dataset with 18,758 samples from 6 categories. Similar to~\cite{mvscn}, the English and French texts are used as two views.

    \item \textbf{CCV20}~\cite{ccv20} contains 6,773 web videos from 20 semantic categories. Here we use two commonly adopted views, \ie, Scale-Invariant Feature Transform (SIFT) visual features and  Mel-Frequency Cepstral Coefficients (MFCC) audio features.

    \item \textbf{HandWritten}~\cite{handwritten} contains 2,000 handwritten digit samples from 10 categories, corresponding to digits 0 through 9. We use Pixel features and Fourier coefficients as two views.
\end{itemize}

\subsection{More Implementation Details}
We implement PLCI in PyTorch and optimize it using Adam~\cite{adam} optimizer with a learning rate of 0.002. The model is trained for 200 epochs with a batch size of 512, of which 20 epochs are used for warmup.
All experiments are conducted on Ubuntu 20.04 with NVIDIA 3090 GPUs. The temperature parameter $\tau_p$ in Eq.~\ref{eq:prototype_assignment} is set to 0.07 across all experiments.
For unavailable counterparts, the transported prototype posterior is sparsified by retaining $k_{\mathrm{top}}$ entries before KL regularization. We set $\alpha=0.2$ and $k_{\mathrm{top}}=3$ across all experiments.

\section{Performance under Only IV or NC}
\label{app:only_iv_or_nc}

Although our proposed method PLCI targets the joint IV+NC setting, we also evaluate it under incomplete view only and noisy correspondence only scenarios. As shown in Tables~\ref{tab:only_nc} and~\ref{tab:only_iv}, PLCI remains competitive when only one challenge exists. Under NC-only settings, PLCI achieves the best average ACC, NMI, and ARI at NCR$=0.2$ and NCR$=0.5$, and still obtains the best average ACC and second-best average NMI and ARI at NCR$=0.8$. Under IV-only settings, PLCI consistently achieves the best average performance across all incomplete view rates. These results show that reliability-aware counterpart rectification and prototype-guided counterpart substitution are both effective in their corresponding single-degradation scenarios, while their unified design further enables PLCI to handle the more practical IV+NC problem.

\begin{table}[htbp]
\centering
\scriptsize
\setlength{\tabcolsep}{2.5pt}
\renewcommand{\arraystretch}{1.1}
\caption{Clustering performance under NC-only settings with IVR fixed to 0. The best and second-best results are shown in \textbf{bold} and \underline{underline}, respectively.}
\label{tab:only_nc}
\resizebox{\textwidth}{!}{%
\begin{tabular}{@{}ll*{18}{c}@{}}
\toprule
\multirow{2}{*}{Setting} & \multicolumn{1}{c}{\multirow{2}{*}{Method}}
& \multicolumn{3}{c}{Scene15}
& \multicolumn{3}{c}{LandUse21}
& \multicolumn{3}{c}{Reuters}
& \multicolumn{3}{c}{CCV20}
& \multicolumn{3}{c}{HandWritten}
& \multicolumn{3}{c}{Average} \\
\cmidrule(lr){3-5}
\cmidrule(lr){6-8}
\cmidrule(lr){9-11}
\cmidrule(lr){12-14}
\cmidrule(lr){15-17}
\cmidrule(lr){18-20}
& & ACC & NMI & ARI & ACC & NMI & ARI & ACC & NMI & ARI & ACC & NMI & ARI & ACC & NMI & ARI & ACC & NMI & ARI \\
\midrule

\multirow{6}{*}{NCR=0.2}
& SURE
& 36.94 & 36.67 & 20.30 & 26.31 & 27.56 & 11.81 & 36.82 & 18.88 & 18.81 & 19.28 & 14.19 & 5.49 & 62.83 & 55.84 & 44.45 & 36.44 & 30.63 & 20.17 \\
& CANDY
& 40.14 & 39.75 & 23.62 & 30.70 & 34.71 & 15.47 & \underline{50.15} & \textbf{32.95} & \underline{25.68} & 20.49 & 16.51 & 6.93 & 88.48 & 79.64 & 77.02 & \underline{45.99} & 40.71 & 29.74 \\
& ROLL
& \textbf{45.87} & \textbf{45.35} & \textbf{28.46} & 28.85 & 32.60 & 14.62 & 46.35 & 22.04 & 21.48 & 21.74 & 16.14 & 6.65 & 80.68 & 66.35 & 62.67 & 44.70 & 36.50 & 26.78 \\
& CorreGen
& 42.94 & 42.94 & 26.41 & \textbf{31.87} & \textbf{38.04} & \textbf{17.64} & 40.36 & 24.74 & 16.58 & \underline{21.81} & \underline{18.43} & \underline{7.34} & \underline{92.31} & \textbf{86.2} & \underline{84.17} & 45.86 & \underline{42.07} & \underline{30.43} \\
& CAMERA
& 42.58 & 40.30 & 23.84 & 26.49 & 28.72 & 12.08 & 49.56 & 27.32 & 23.44 & 20.34 & 15.83 & 6.26 & 79.55 & 64.89 & 62.61 & 43.70 & 35.41 & 25.65 \\
& Ours
& \underline{44.92} & \underline{44.92} & \underline{28.10} & \underline{30.99} & \underline{35.13} & \underline{16.52} & \textbf{53.02} & \underline{32.44} & \textbf{26.69} & \textbf{24.46} & \textbf{19.64} & \textbf{8.40} & \textbf{92.58} & \underline{84.86} & \textbf{84.27} & \textbf{49.19} & \textbf{43.40} & \textbf{32.80} \\

\midrule
\multirow{6}{*}{NCR=0.5}
& SURE
& 36.04 & 35.92 & 20.04 & 23.42 & 23.49 & 9.07 & 38.34 & 16.70 & 13.20 & 15.50 & 11.10 & 3.74 & 61.32 & 52.66 & 41.58 & 34.92 & 27.97 & 17.53 \\
& CANDY
& 40.54 & 39.04 & 23.62 & 29.56 & 32.93 & 14.84 & 43.79 & \underline{24.91} & 19.04 & 18.24 & 14.30 & 5.49 & 83.06 & 74.29 & 69.57 & 43.04 & 37.09 & 26.51 \\
& ROLL
& 41.27 & 36.32 & 22.62 & 28.75 & 32.63 & 14.42 & \underline{45.72} & 20.78 & \underline{20.77} & 18.33 & 10.93 & 4.22 & 82.57 & 69.63 & 65.90 & 43.33 & 34.06 & 25.59 \\
& CorreGen
& \textbf{42.37} & \underline{41.10} & \textbf{24.94} & \textbf{31.99} & \textbf{37.95} & \textbf{18.17} & 40.22 & 21.80 & 14.51 & \underline{21.78} & \underline{18.64} & \underline{7.43} & \underline{88.86} & \textbf{83.67} & \underline{80.26} & \underline{45.04} & \underline{40.63} & \underline{29.06} \\
& CAMERA
& 34.40 & 34.88 & 18.33 & 20.53 & 30.25 & 6.89 & 43.86 & 18.36 & 15.64 & 17.93 & 13.07 & 4.84 & 63.94 & 40.73 & 37.11 & 36.13 & 27.46 & 16.56 \\
& Ours
& \underline{42.27} & \textbf{41.26} & \underline{24.39} & \underline{31.40} & \underline{35.74} & \underline{17.01} & \textbf{49.70} & \textbf{27.84} & \textbf{23.35} & \textbf{23.64} & \textbf{19.37} & \textbf{8.02} & \textbf{91.13} & \underline{82.42} & \textbf{81.47} & \textbf{47.63} & \textbf{41.33} & \textbf{30.85} \\

\midrule
\multirow{6}{*}{NCR=0.8}
& SURE
& 26.40 & 28.61 & 13.64 & 18.59 & 18.15 & 6.01 & 36.31 & 16.29 & 13.40 & 12.41 & 9.56 & 2.79 & 60.50 & 51.46 & 40.00 & 30.84 & 24.81 & 15.17 \\
& CANDY
& 39.24 & 36.54 & 21.50 & 27.61 & 30.52 & 13.22 & \underline{39.45} & 15.93 & 13.71 & 17.56 & 12.46 & 4.39 & 53.98 & 48.05 & 34.46 & 35.57 & 28.70 & 17.46 \\
& ROLL
& 31.23 & 21.24 & 12.58 & 26.93 & 29.23 & 12.92 & 38.28 & 16.12 & \underline{14.94} & 11.57 & 3.90 & 0.40 & 76.92 & \underline{69.78} & 62.31 & 36.99 & 28.05 & 20.63 \\
& CorreGen
& \underline{39.42} & \textbf{38.64} & \underline{22.50} & \textbf{31.71} & \textbf{37.50} & \textbf{17.67} & 35.72 & \underline{16.97} & 9.76 & \underline{21.03} & \textbf{17.88} & \underline{6.99} & \textbf{85.42} & \textbf{77.89} & \textbf{73.68} & \underline{42.66} & \textbf{37.78} & \textbf{26.12} \\
& CAMERA
& 34.62 & 33.93 & 18.13 & 15.32 & 13.73 & 3.56 & 34.67 & 11.59 & 9.49 & 14.97 & 10.27 & 3.47 & 33.29 & 12.79 & 9.20 & 26.57 & 16.46 & 8.77 \\
& Ours
& \textbf{40.25} & \underline{38.36} & \textbf{22.59} & \underline{29.58} & \underline{33.38} & \underline{15.43} & \textbf{43.34} & \textbf{21.60} & \textbf{17.16} & \textbf{21.82} & \underline{17.06} & \textbf{7.19} & \underline{79.03} & 69.26 & \underline{64.15} & \textbf{42.80} & \underline{35.93} & \underline{25.30} \\
\bottomrule
\end{tabular}%
}
\end{table}

\begin{table}[htbp]
\centering
\scriptsize
\setlength{\tabcolsep}{2.5pt}
\renewcommand{\arraystretch}{1.1}
\caption{Clustering performance under IV-only settings with NCR fixed to 0. The best and second-best results are shown in \textbf{bold} and \underline{underline}, respectively.}
\label{tab:only_iv}
\resizebox{\textwidth}{!}{%
\begin{tabular}{@{}ll*{18}{c}@{}}
\toprule
\multirow{2}{*}{Setting} & \multicolumn{1}{c}{\multirow{2}{*}{Method}}
& \multicolumn{3}{c}{Scene15}
& \multicolumn{3}{c}{LandUse21}
& \multicolumn{3}{c}{Reuters}
& \multicolumn{3}{c}{CCV20}
& \multicolumn{3}{c}{HandWritten}
& \multicolumn{3}{c}{Average} \\
\cmidrule(lr){3-5}
\cmidrule(lr){6-8}
\cmidrule(lr){9-11}
\cmidrule(lr){12-14}
\cmidrule(lr){15-17}
\cmidrule(lr){18-20}
& & ACC & NMI & ARI & ACC & NMI & ARI & ACC & NMI & ARI & ACC & NMI & ARI & ACC & NMI & ARI & ACC & NMI & ARI \\
\midrule

\multirow{6}{*}{IVR=0.2}
& SURE
& 41.16 & 42.51 & 24.79 & 26.13 & 30.00 & 12.48 & 45.65 & 27.65 & 21.16 & 19.45 & 16.52 & 6.16 & 69.01 & 61.33 & 52.83 & 40.28 & 35.60 & 23.48 \\
& GCFAgg
& 39.73 & 40.46 & 22.57 & 25.70 & 29.05 & 12.12 & 31.69 & 14.68 & 5.76 & 18.31 & 15.30 & 5.41 & 86.65 & 76.61 & 72.74 & 40.42 & 35.22 & 23.72 \\
& DIVIDE
& \textbf{46.39} & \textbf{46.27} & \textbf{29.43} & \underline{30.82} & \textbf{37.51} & \underline{16.93} & \underline{53.73} & \textbf{36.70} & 27.59 & \underline{23.02} & 17.60 & 7.29 & \underline{92.73} & \underline{85.55} & \underline{84.77} & \underline{49.34} & \underline{44.73} & \underline{33.20} \\
& FreeCSL
& 40.25 & 43.27 & 24.81 & 28.77 & 34.07 & 15.32 & 49.52 & 30.89 & 27.53 & 21.13 & 17.36 & 6.89 & 85.64 & 82.68 & 77.56 & 45.06 & 41.65 & 30.42 \\
& CAMERA
& \underline{45.98} & 45.73 & 27.91 & 28.18 & 34.18 & 14.53 & 54.92 & 34.66 & \underline{29.86} & 22.50 & \underline{19.07} & \underline{7.69} & 89.37 & 83.83 & 80.39 & 48.19 & 43.49 & 32.08 \\
& Ours
& 45.17 & \underline{46.00} & \underline{29.21} & \textbf{32.13} & \underline{37.26} & \textbf{18.11} & \textbf{58.36} & \underline{36.02} & \textbf{31.83} & \textbf{25.37} & \textbf{21.11} & \textbf{9.22} & \textbf{93.72} & \textbf{87.88} & \textbf{86.67} & \textbf{50.95} & \textbf{45.65} & \textbf{35.01} \\

\midrule
\multirow{6}{*}{IVR=0.5}
& SURE
& 39.45 & 41.32 & 23.32 & 25.19 & 28.45 & 11.88 & 46.31 & 27.87 & 22.90 & 19.64 & 16.24 & 6.07 & 68.07 & 60.53 & 51.65 & 39.73 & 34.88 & 23.16 \\
& GCFAgg
& 36.72 & 38.49 & 20.96 & 24.23 & 26.51 & 10.64 & 34.46 & 16.46 & 6.74 & 18.00 & 14.83 & 5.22 & 77.86 & 70.24 & 62.58 & 38.25 & 33.31 & 21.23 \\
& DIVIDE
& \underline{44.08} & \textbf{44.57} & \textbf{27.59} & \underline{30.33} & \underline{35.54} & \underline{16.25} & 49.77 & \textbf{35.93} & 26.30 & \underline{22.15} & 17.27 & 6.98 & \underline{90.28} & \underline{81.88} & 80.24 & \underline{47.32} & \underline{43.04} & \underline{31.47} \\
& FreeCSL
& 39.39 & 40.86 & 23.06 & 27.76 & 32.21 & 14.13 & 47.20 & 29.03 & 24.77 & 20.01 & 16.56 & 6.33 & 77.12 & 78.08 & \underline{80.62} & 42.30 & 39.35 & 29.78 \\
& CAMERA
& \textbf{44.85} & 43.73 & 26.42 & 26.81 & 31.85 & 13.22 & \textbf{53.43} & 32.84 & \underline{28.35} & 21.98 & \underline{18.17} & \underline{7.46} & 87.52 & 79.51 & 76.07 & 46.92 & 41.22 & 30.30 \\
& Ours
& 42.50 & \underline{44.55} & \underline{27.13} & \textbf{30.64} & \textbf{37.39} & \textbf{17.22} & \underline{53.21} & \underline{33.01} & \textbf{28.94} & \textbf{24.97} & \textbf{21.33} & \textbf{9.16} & \textbf{92.99} & \textbf{86.75} & \textbf{85.21} & \textbf{48.86} & \textbf{44.61} & \textbf{33.53} \\

\midrule
\multirow{6}{*}{IVR=0.8}
& SURE
& 38.85 & 38.05 & 21.94 & 22.03 & 23.28 & 9.13 & 48.71 & 31.21 & 25.33 & 19.15 & 15.14 & 5.73 & 61.89 & 54.40 & 44.28 & 38.13 & 32.42 & 21.28 \\
& GCFAgg
& 38.68 & 37.46 & 21.98 & 18.20 & 18.41 & 5.95 & 35.01 & 19.29 & 8.47 & 16.13 & 12.29 & 4.11 & 59.99 & 54.35 & 38.64 & 33.60 & 28.36 & 15.83 \\
& DIVIDE
& \textbf{43.12} & \textbf{42.88} & \textbf{26.49} & \underline{29.1} & \underline{32.65} & \underline{14.88} & \textbf{53.37} & \textbf{34.82} & \textbf{30.52} & \underline{21.01} & \underline{16.22} & 6.06 & \underline{84.4} & 73.42 & \underline{70.07} & \underline{46.20} & \underline{40.00} & \underline{29.60} \\
& FreeCSL
& 38.08 & 38.06 & 21.89 & 26.12 & 28.46 & 12.19 & 46.94 & 28.48 & 24.73 & 20.31 & 15.87 & 6.26 & 74.00 & \underline{79.56} & 62.31 & 41.09 & 38.09 & 25.48 \\
& CAMERA
& 40.33 & 40.42 & 22.68 & 22.97 & 27.49 & 9.74 & \underline{51.79} & \underline{32.44} & \underline{27.13} & 19.51 & 16.18 & \underline{6.35} & 83.22 & 71.59 & 67.73 & 43.56 & 37.62 & 26.73 \\
& Ours
& \underline{41.26} & \underline{42.86} & \underline{26.17} & \textbf{30.62} & \textbf{37.80} & \textbf{16.54} & 48.56 & 28.38 & 24.95 & \textbf{24.03} & \textbf{20.57} & \textbf{8.31} & \textbf{91.58} & \textbf{84.39} & \textbf{82.41} & \textbf{47.21} & \textbf{42.80} & \textbf{31.68} \\
\bottomrule
\end{tabular}%
}
\end{table}

\section{Limitations of Two-Stage Combination of IV- and NC-Oriented Methods}
\label{sec:combine}
A straightforward solution to imperfect information is to combine an IV-oriented method and an NC-oriented method. To examine this strategy, we sequentially combine DIVIDE and CorreGen in two orders. D+C means that it first uses DIVIDE to recover the incomplete views and then applies CorreGen to re-align the recovered multi-view data for clustering. Conversely, C+D first uses CorreGen to correct noisy correspondences and then applies DIVIDE to recover the remaining missing views. The results are reported under two settings, \ie, IIR=0.5 and IIR=0.8. 

As shown in Table~\ref{tab:combine_order}, sequentially combining IV- and NC-oriented methods does not provide a reliable solution to imperfect information. Under IIR=0.5, C+D generally outperforms D+C, suggesting that correcting noisy correspondences before incomplete view recovery is less harmful in the moderate setting. However, C+D is only competitive on Scene15 and remains clearly inferior to PLCI on LandUse21, CCV20, and HandWritten. Under the more severe setting, \ie, IIR=0.8, both two-stage combinations degrade substantially, and their relative performance becomes order-dependent. In contrast, PLCI achieves the best results on most datasets and metrics. These results indicate that incomplete view recovery and correspondence re-alignment are tightly coupled, and simply cascading existing IV- and NC-oriented methods cannot effectively handle unavailable and unreliable counterparts simultaneously.

The reason is that the two subproblems are tightly coupled. View recovery requires reliable cross-view correspondences, while correspondence correction requires sufficiently complete instances. In a two-stage pipeline, errors produced in the first stage are propagated to the second stage, leading to suboptimal clustering. In contrast, PLCI models unavailable and unreliable counterparts in a unified latent counterpart inference framework, which avoids the error accumulation of sequential recovery and correction.

\begin{table*}[htbp]
\centering
\scriptsize
\setlength{\tabcolsep}{3.0pt}
\renewcommand{\arraystretch}{1.08}
\caption{Comparison of two-stage combinations and PLCI under joint IV and NC. C+D denotes applying DIVIDE after CorreGen, while D+C denotes applying CorreGen after DIVIDE.}
\label{tab:combine_order}
\resizebox{\textwidth}{!}{%
\begin{tabular}{@{}ll*{12}{c}@{}}
\toprule
\multirow{2}{*}{Setting} & \multirow{2}{*}{Method}
& \multicolumn{3}{c}{Scene15}
& \multicolumn{3}{c}{LandUse21}
& \multicolumn{3}{c}{CCV20}
& \multicolumn{3}{c}{HandWritten} \\
\cmidrule(lr){3-5}
\cmidrule(lr){6-8}
\cmidrule(lr){9-11}
\cmidrule(lr){12-14}
& & ACC & NMI & ARI & ACC & NMI & ARI & ACC & NMI & ARI & ACC & NMI & ARI \\
\midrule
\multirow{3}{*}{IIR=0.5}
& C+D
& 41.58 & 41.34 & \textbf{25.05}
& 29.51 & 33.12 & 14.80
& 19.33 & 14.98 & 5.87
& 85.66 & 73.48 & 71.57 \\
& D+C
& 38.63 & 35.14 & 20.57
& 26.50 & 27.78 & 12.29
& 16.39 & 9.54 & 3.45
& 76.71 & 59.86 & 57.09 \\
& Ours
& \multicolumn{1}{>{\columncolor{gray!20}}c}{\textbf{42.17}}
& \multicolumn{1}{>{\columncolor{gray!20}}c}{\textbf{41.39}}
& \multicolumn{1}{>{\columncolor{gray!20}}c}{25.00}
& \multicolumn{1}{>{\columncolor{gray!20}}c}{\textbf{30.09}}
& \multicolumn{1}{>{\columncolor{gray!20}}c}{\textbf{34.58}}
& \multicolumn{1}{>{\columncolor{gray!20}}c}{\textbf{16.13}}
& \multicolumn{1}{>{\columncolor{gray!20}}c}{\textbf{22.97}}
& \multicolumn{1}{>{\columncolor{gray!20}}c}{\textbf{18.78}}
& \multicolumn{1}{>{\columncolor{gray!20}}c}{\textbf{8.06}}
& \multicolumn{1}{>{\columncolor{gray!20}}c}{\textbf{86.94}}
& \multicolumn{1}{>{\columncolor{gray!20}}c}{\textbf{76.08}}
& \multicolumn{1}{>{\columncolor{gray!20}}c}{\textbf{74.05}} \\
\midrule
\multirow{3}{*}{IIR=0.8}
& C+D
& 30.47 & 22.22 & 12.23
& 20.34 & 18.99 & 6.54
& 11.37 & 5.11 & 1.43
& 49.03 & 31.04 & 23.89 \\
& D+C
& 26.71 & 31.16 & 14.58
& 21.33 & 23.31 & 8.26
& 16.29 & 10.78 & 3.69
& 45.55 & 41.48 & 27.54 \\
& Ours
& \multicolumn{1}{>{\columncolor{gray!20}}c}{\textbf{37.78}}
& \multicolumn{1}{>{\columncolor{gray!20}}c}{\textbf{36.71}}
& \multicolumn{1}{>{\columncolor{gray!20}}c}{\textbf{22.21}}
& \multicolumn{1}{>{\columncolor{gray!20}}c}{\textbf{26.66}}
& \multicolumn{1}{>{\columncolor{gray!20}}c}{\textbf{30.68}}
& \multicolumn{1}{>{\columncolor{gray!20}}c}{\textbf{11.68}}
& \multicolumn{1}{>{\columncolor{gray!20}}c}{\textbf{16.47}}
& \multicolumn{1}{>{\columncolor{gray!20}}c}{\textbf{12.39}}
& \multicolumn{1}{>{\columncolor{gray!20}}c}{\textbf{4.35}}
& \multicolumn{1}{>{\columncolor{gray!20}}c}{\textbf{72.54}}
& \multicolumn{1}{>{\columncolor{gray!20}}c}{\textbf{63.07}}
& \multicolumn{1}{>{\columncolor{gray!20}}c}{\textbf{57.13}} \\
\bottomrule
\end{tabular}%
}
\end{table*}

\section{Performance under Different IVR and NCR}
\label{sec:different_ivr_ncr}

\begin{figure}[htbp]
  \centering
  \includegraphics[width=\linewidth]{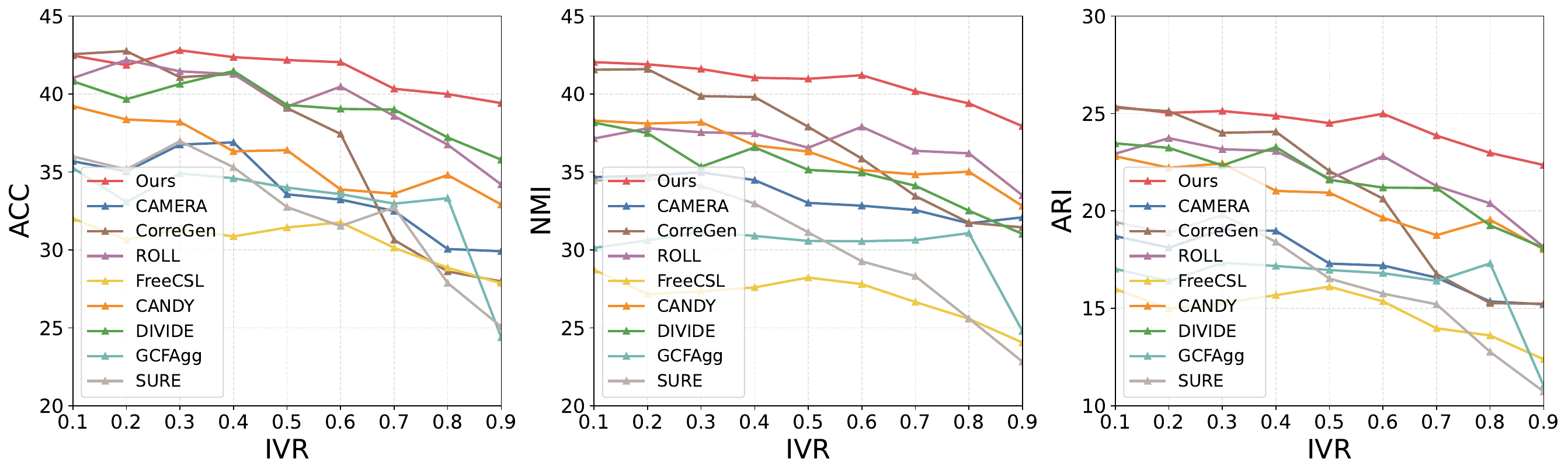} 
  \caption{Clustering performance under different IVR with a fixed NCR=0.5 on Scene15.}
  \label{fig:scene15_fixed_fp}
\end{figure}
\begin{figure}[htbp]
  \centering
  \includegraphics[width=\linewidth]{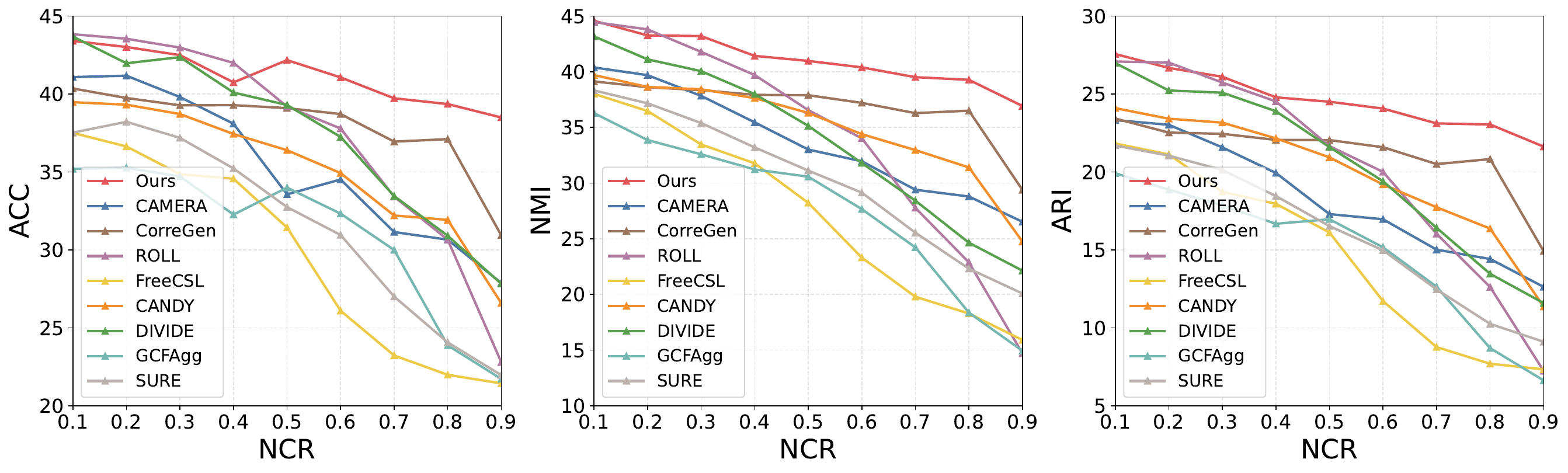} 
  \caption{Clustering performance under different NCR with a fixed IVR=0.5 on Scene15.}
  \label{fig:scene15_fixed_mr}
\end{figure}

To further verify the robustness of PLCI, we conduct additional experiments by varying one degradation rate from 0.1 to 0.9 with an interval of 0.1 while fixing the other at 0.5. Specifically, we first vary IVR while fixing NCR at 0.5, and then vary NCR while fixing IVR at 0.5. These experiments provide a more fine-grained robustness analysis beyond the paired NCR/IVR settings reported in Table~\ref{tab:main_results}.

Fig.~\ref{fig:scene15_fixed_fp} reports the results on Scene15 under different IVR settings with NCR fixed at 0.5. PLCI maintains highly competitive performance across almost all IVR settings in terms of ACC, NMI, and ARI, whereas most baselines exhibit clear performance degradation as IVR increases. This indicates that PLCI is less sensitive to the reduction of available cross-view counterparts, benefiting from its prototype-level semantic transport for unavailable counterparts. Fig.~\ref{fig:scene15_fixed_mr} shows the results on Scene15 under different NCR settings with IVR fixed at 0.5. When NCR is small, PLCI already performs competitively with the strongest baselines. As NCR increases, PLCI becomes consistently superior, while most competing methods degrade sharply. This demonstrates that PLCI can effectively mitigate the negative effect of unreliable counterparts by adaptively reducing the reliance on corrupted observed counterparts.

To further examine the generality of this robustness, we provide additional results on LandUse21, Reuters, CCV20, and HandWritten under the same settings. The results are shown in Fig.~\ref{fig:landuse21_robustness}, Fig.~\ref{fig:reuters_robustness}, Fig.~\ref{fig:ccv20_robustness}, and Fig.~\ref{fig:handwritten_robustness}. We observe that PLCI is consistently more robust than baselines across different datasets and degradation settings. Moreover, the performance gap between PLCI and competing methods generally widens as the degradation rate increases, indicating that PLCI is particularly effective in handling severe imperfect information.

Overall, these experiments verify the robustness of PLCI under both increasing IVR and increasing NCR. The stable performance further demonstrates the effectiveness of inferring latent counterparts for both unavailable and unreliable cross-view counterparts.

\begin{figure*}[htbp]
  \centering
  \includegraphics[width=0.9\linewidth]{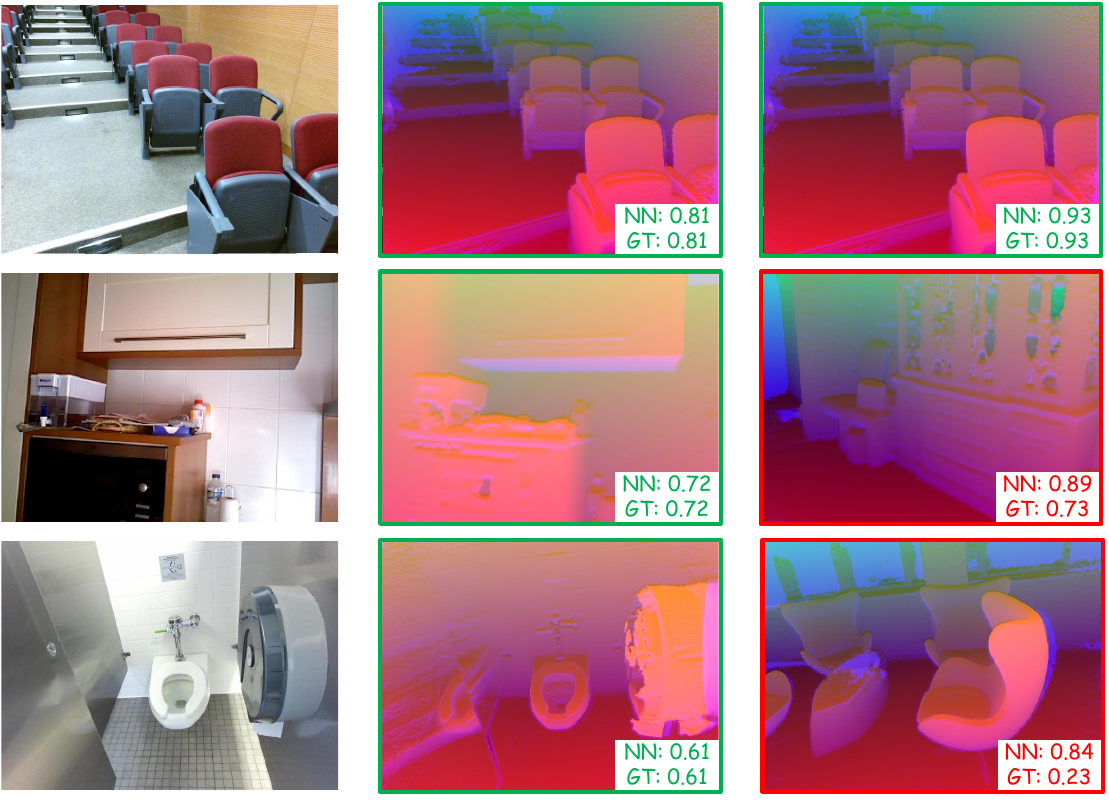} 
  \caption{Case study on SUN RGB-D under IIR$=0.5$. The first column shows the observed RGB views. The second and third columns show the depth maps whose features have the highest cosine similarity to the recovered depth-view features produced by PLCI and DIVIDE, respectively. From top to bottom, the ground-truth categories are \textit{lecture theatre}, \textit{dining room}, and \textit{bathroom}. NN denotes the cosine similarity between the recovered depth-view feature and its nearest-neighbor depth feature, while GT denotes the cosine similarity between the recovered depth-view feature and the ground-truth depth counterpart.}

  \label{fig:case_study}
\end{figure*}

\section{Case study}
\label{case_study}
We conduct a case study on SUN RGB-D under IIR$=0.5$. Given an observed RGB image, each method recovers the missing depth-view feature. We then retrieve the depth map whose feature is most similar to the recovered feature. As shown in Fig.~\ref{fig:case_study}, PLCI retrieves the ground-truth depth counterpart in all cases. By contrast, DIVIDE succeeds only in the first case, retrieving a same-category but incorrect depth map for \textit{dining room} and even a wrong-category depth map for \textit{bathroom}. These results show that PLCI recovers more reliable missing-view counterparts under imperfect information.

\begin{figure*}[htbp]
  \centering
  \begin{subfigure}{0.9\textwidth}
    \centering
    \includegraphics[width=\linewidth]{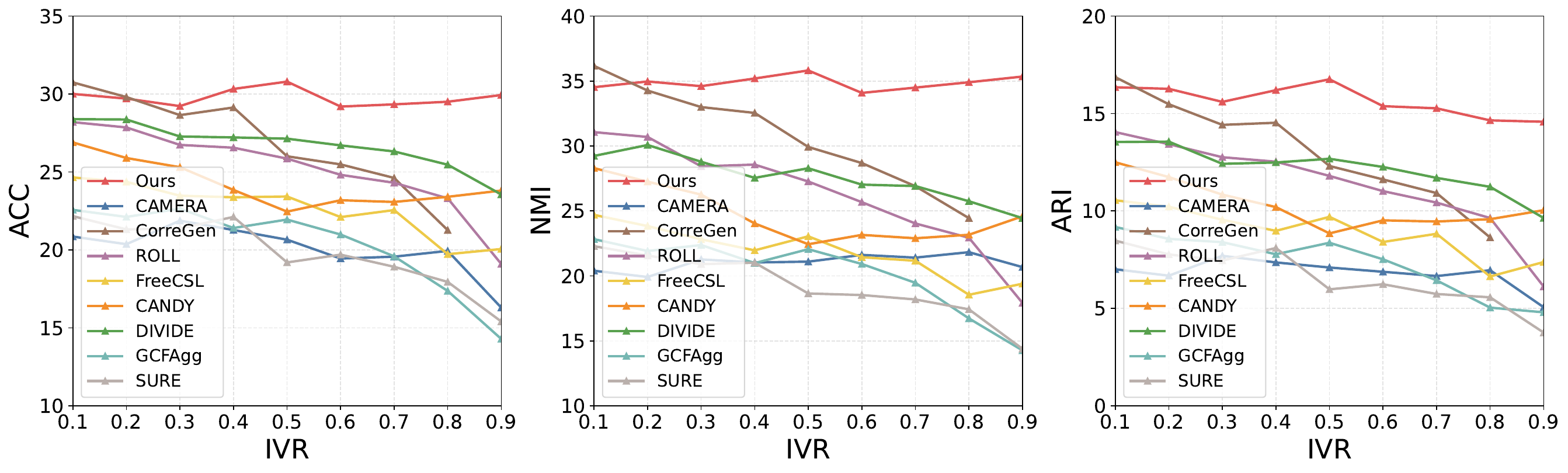}
    \caption{Different IVR with NCR fixed at 0.5.}
    \label{fig:landuse21_fixed_ncr}
  \end{subfigure}

  \vspace{0.8em}

  \begin{subfigure}{0.9\textwidth}
    \centering
    \includegraphics[width=\linewidth]{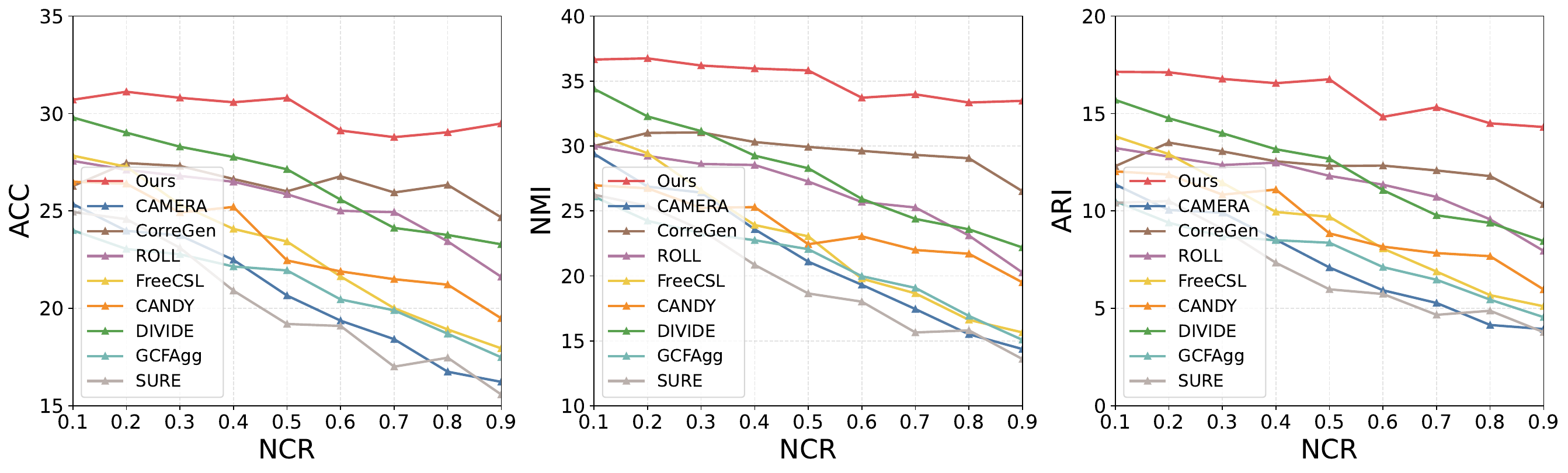}
    \caption{Different NCR with IVR fixed at 0.5.}
    \label{fig:landuse21_fixed_ivr}
  \end{subfigure}

  \caption{Additional robustness analysis on LandUse21 under different IVR and NCR settings.}
  \label{fig:landuse21_robustness}
\end{figure*}

\begin{figure*}[htbp]
  \centering
  \begin{subfigure}{0.9\textwidth}
    \centering
    \includegraphics[width=\linewidth]{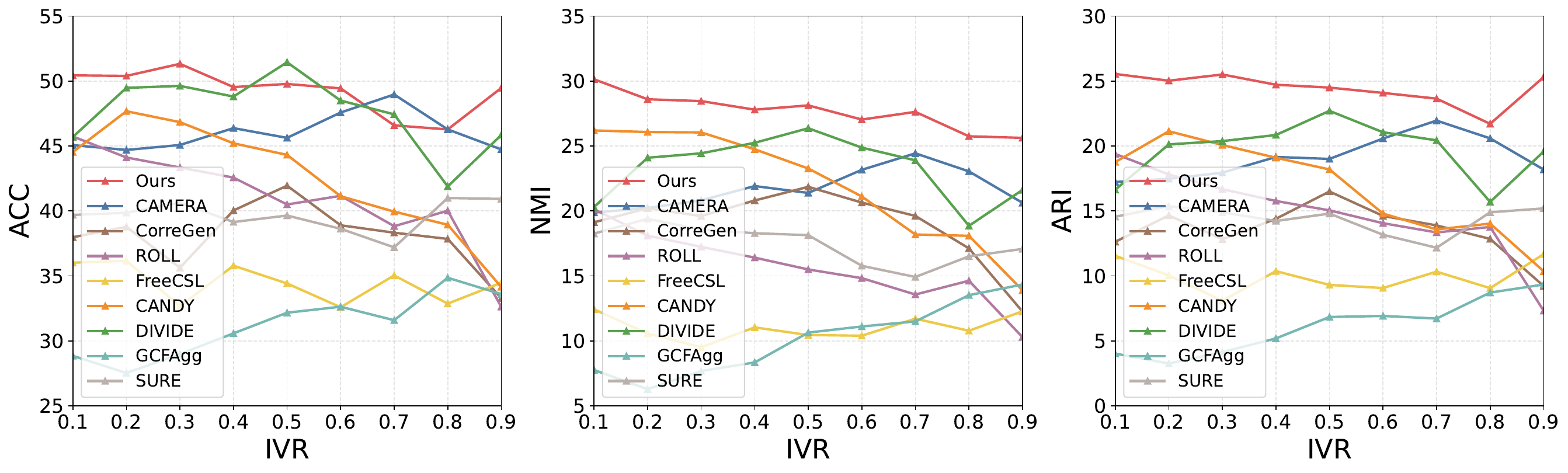}
    \caption{Different IVR with NCR fixed at 0.5.}
    \label{fig:reuters_fixed_ncr}
  \end{subfigure}

  \vspace{0.8em}

  \begin{subfigure}{0.9\textwidth}
    \centering
    \includegraphics[width=\linewidth]{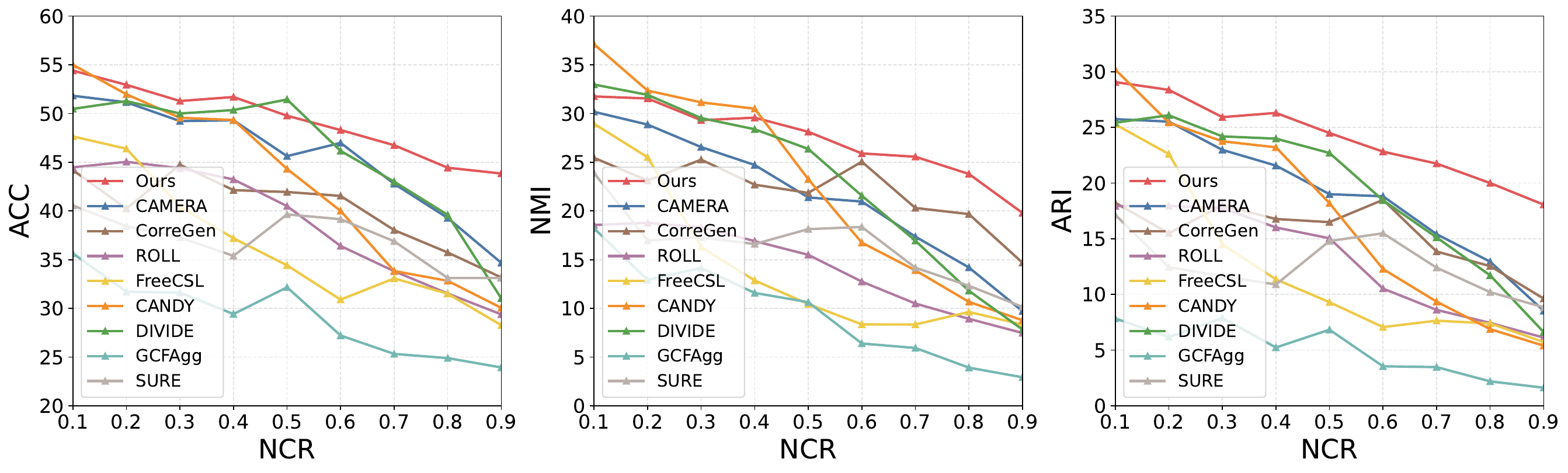}
    \caption{Different NCR with IVR fixed at 0.5.}
    \label{fig:reuters_fixed_ivr}
  \end{subfigure}

  \caption{Additional robustness analysis on Reuters under different IVR and NCR settings.}
  \label{fig:reuters_robustness}
\end{figure*}

\begin{figure*}[htbp]
  \centering
  \begin{subfigure}{0.9\textwidth}
    \centering
    \includegraphics[width=\linewidth]{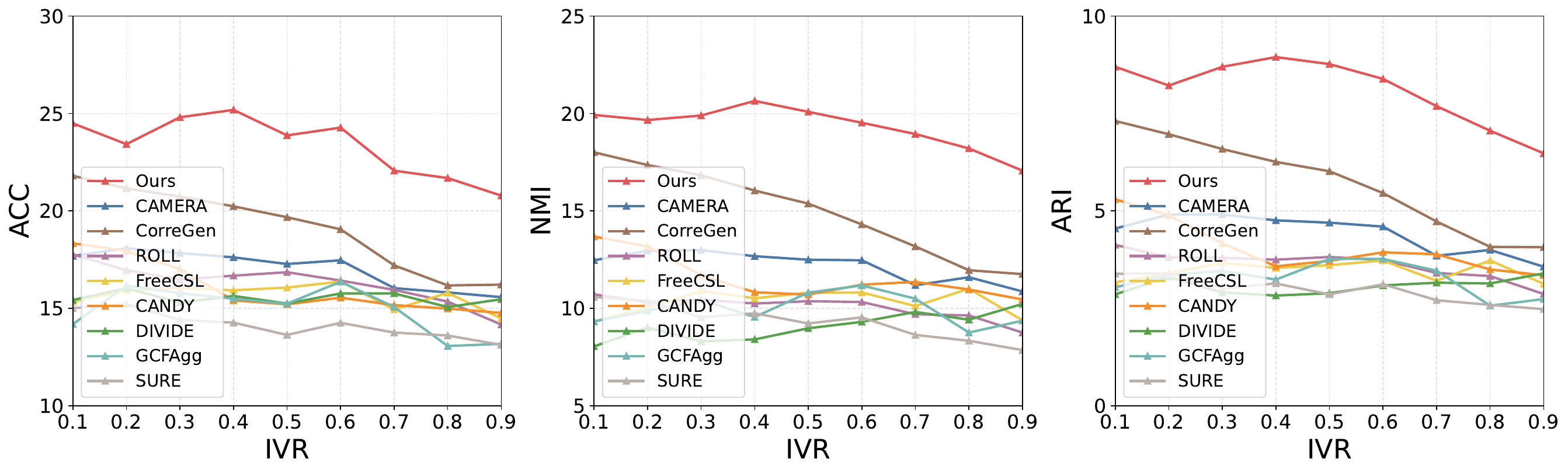}
    \caption{Different IVR with NCR fixed at 0.5.}
    \label{fig:ccv20_fixed_ncr}
  \end{subfigure}

  \vspace{0.8em}

  \begin{subfigure}{0.9\textwidth}
    \centering
    \includegraphics[width=\linewidth]{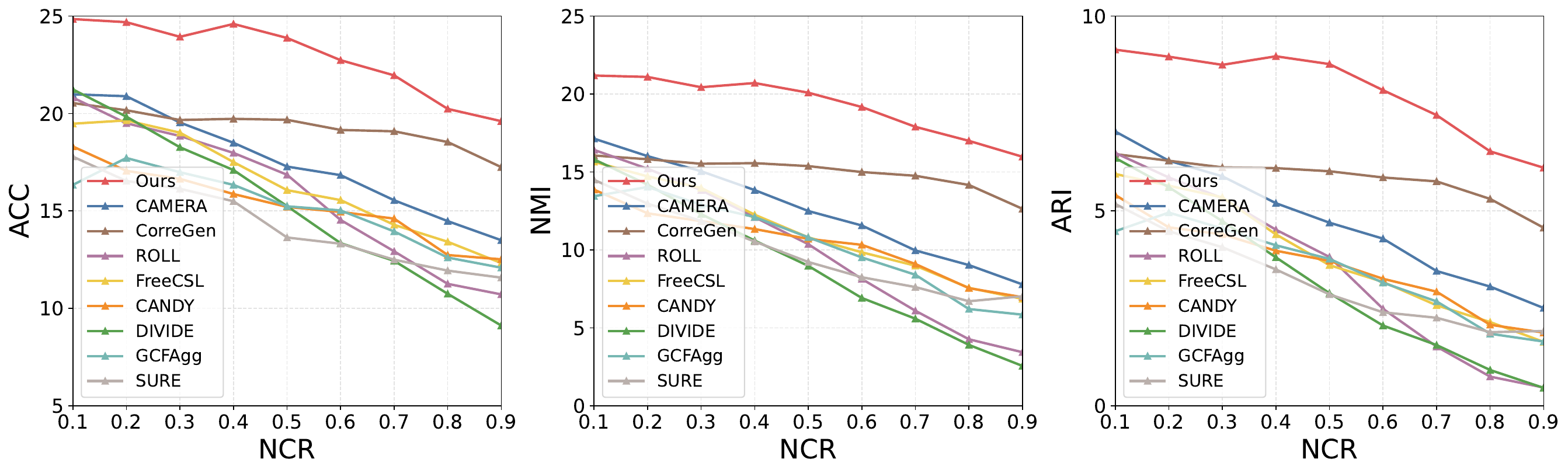}
    \caption{Different NCR with IVR fixed at 0.5.}
    \label{fig:ccv20_fixed_ivr}
  \end{subfigure}

  \caption{Additional robustness analysis on CCV20 under different IVR and NCR settings.}
  \label{fig:ccv20_robustness}
\end{figure*}

\begin{figure*}[htbp]
  \centering
  \begin{subfigure}{0.9\textwidth}
    \centering
    \includegraphics[width=\linewidth]{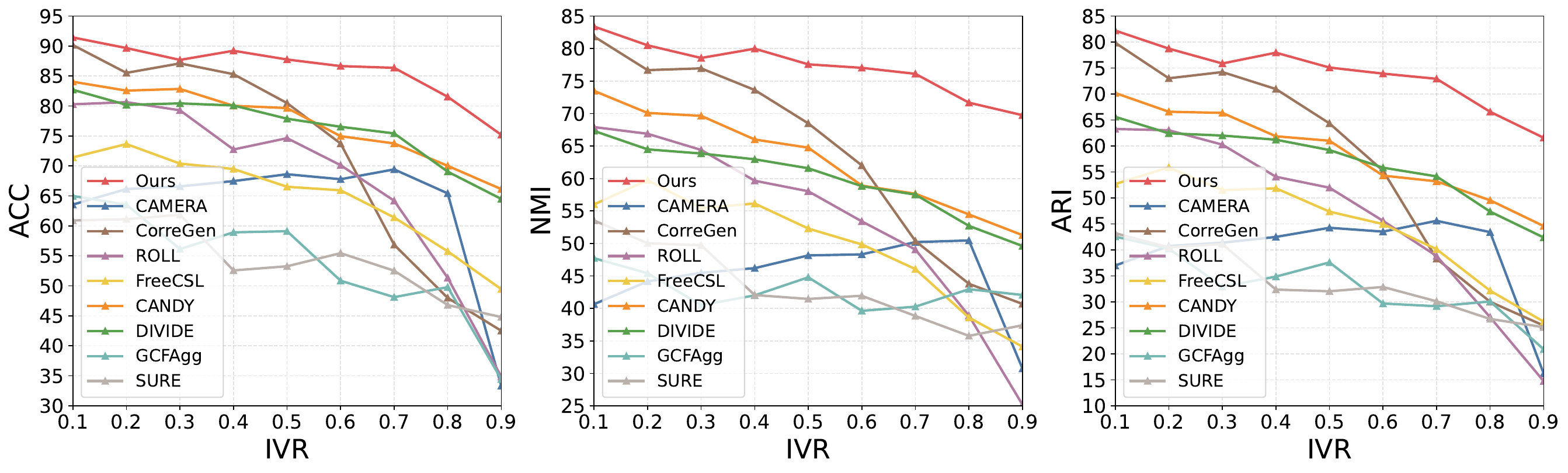}
    \caption{Different IVR with NCR fixed at 0.5.}
    \label{fig:handwritten_fixed_ncr}
  \end{subfigure}

  \vspace{0.8em}

  \begin{subfigure}{0.9\textwidth}
    \centering
    \includegraphics[width=\linewidth]{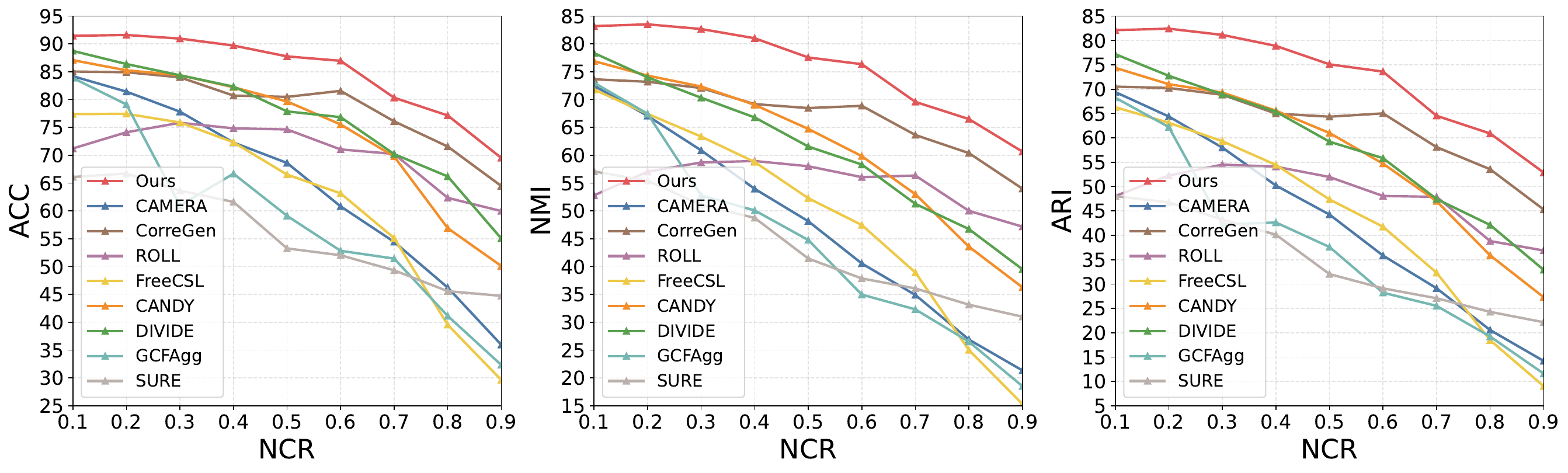}
    \caption{Different NCR with IVR fixed at 0.5.}
    \label{fig:handwritten_fixed_ivr}
  \end{subfigure}

  \caption{Additional robustness analysis on HandWritten under different IVR and NCR settings.}
  \label{fig:handwritten_robustness}
\end{figure*}

\section{Visualization}
\begin{figure}[htbp]
    \centering
        \begin{subfigure}[htbp]{0.24\linewidth}
        \centering
        \includegraphics[width=\linewidth]{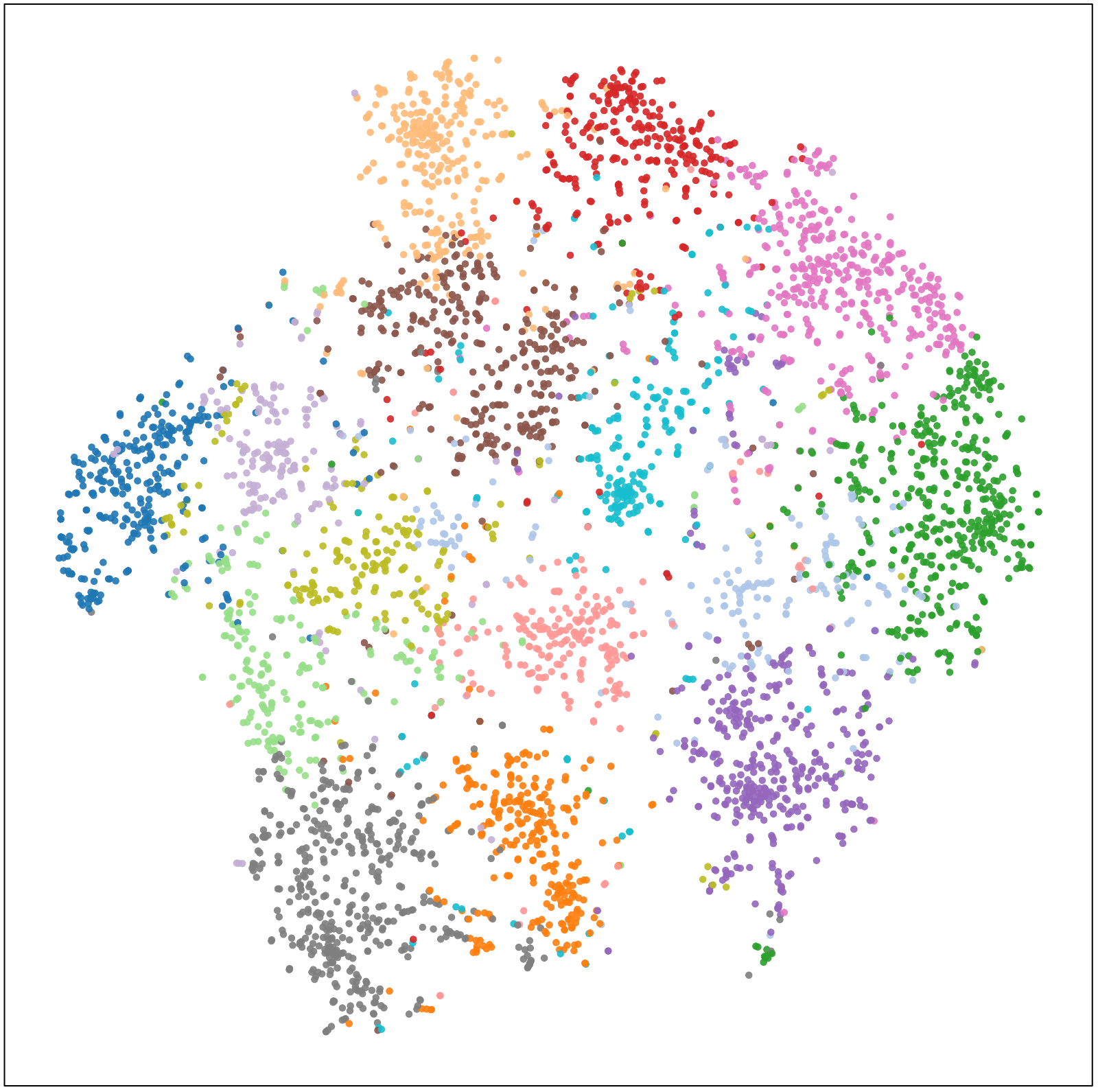}
        \caption{DIVIDE}
        \label{fig:tsne_divide}
    \end{subfigure}
    \hfill
    \begin{subfigure}[htbp]{0.24\linewidth}
        \centering
        \includegraphics[width=\linewidth]{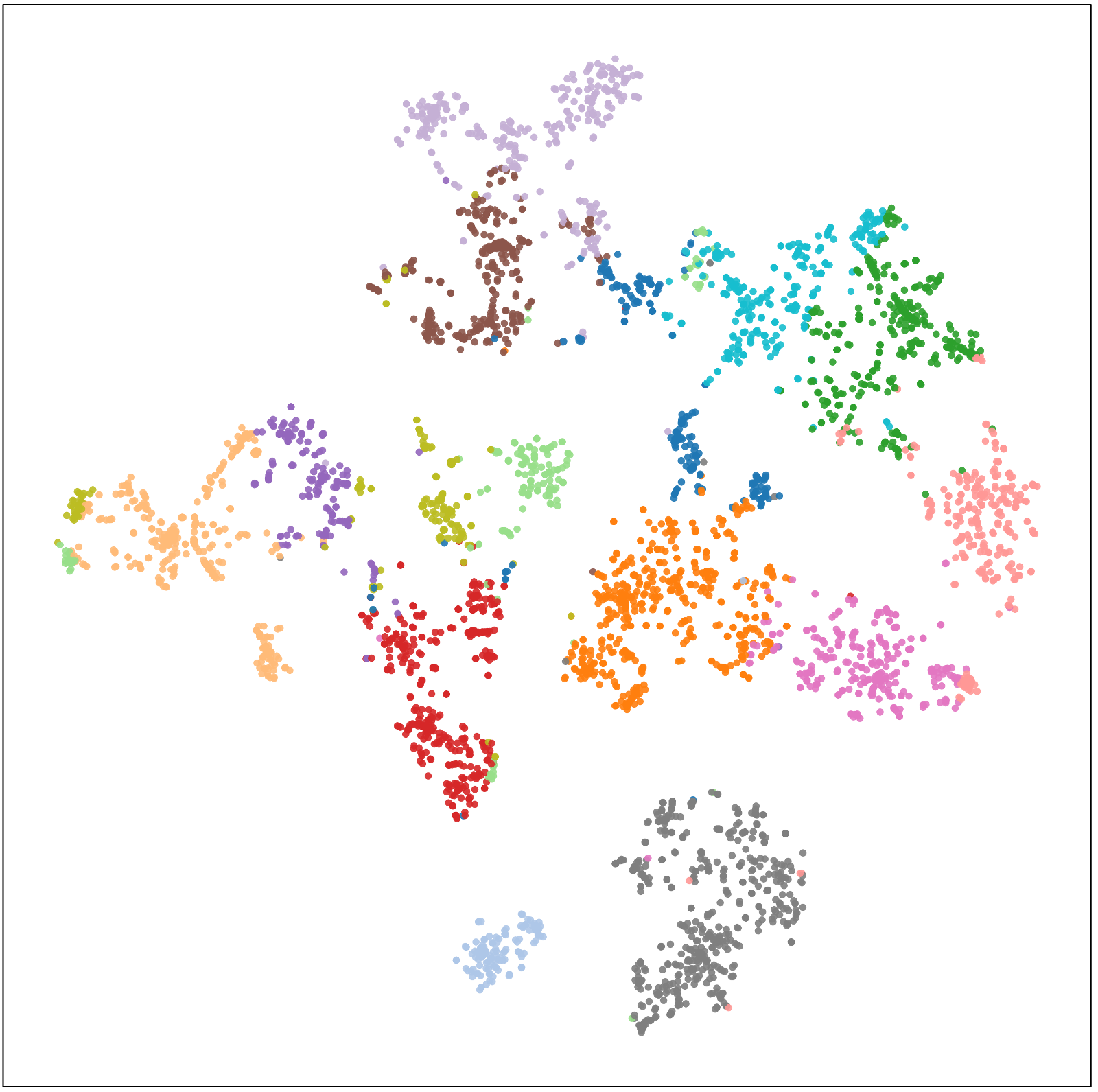}
        \caption{CANDY}
        \label{fig:tsne_candy}
    \end{subfigure}
    \hfill
    \begin{subfigure}[htbp]{0.24\linewidth}
        \centering
        \includegraphics[width=\linewidth]{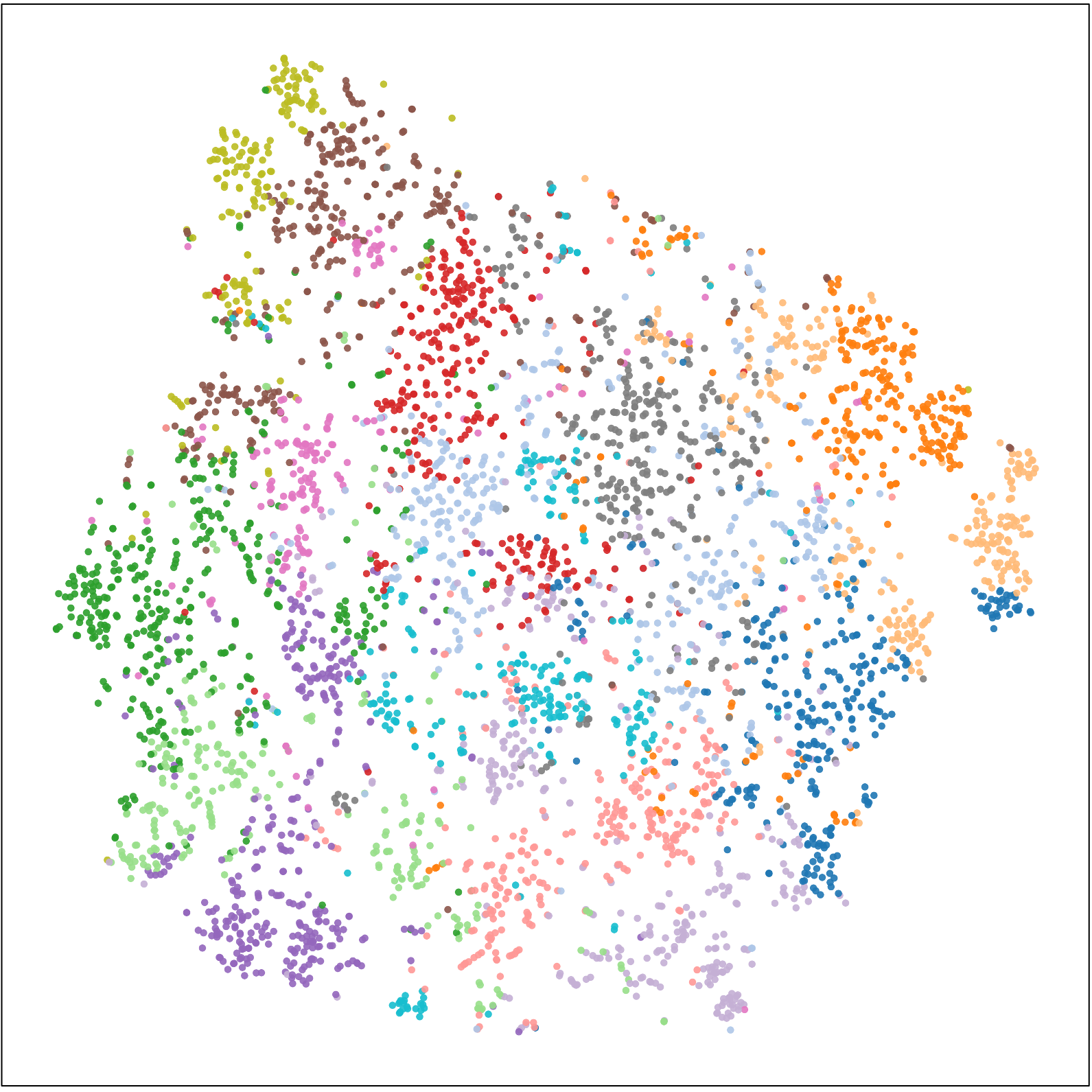}
        \caption{CAMERA}
        \label{fig:tsne_camera}
    \end{subfigure}
    \hfill
    \begin{subfigure}[htbp]{0.24\linewidth}
        \centering
        \includegraphics[width=\linewidth]{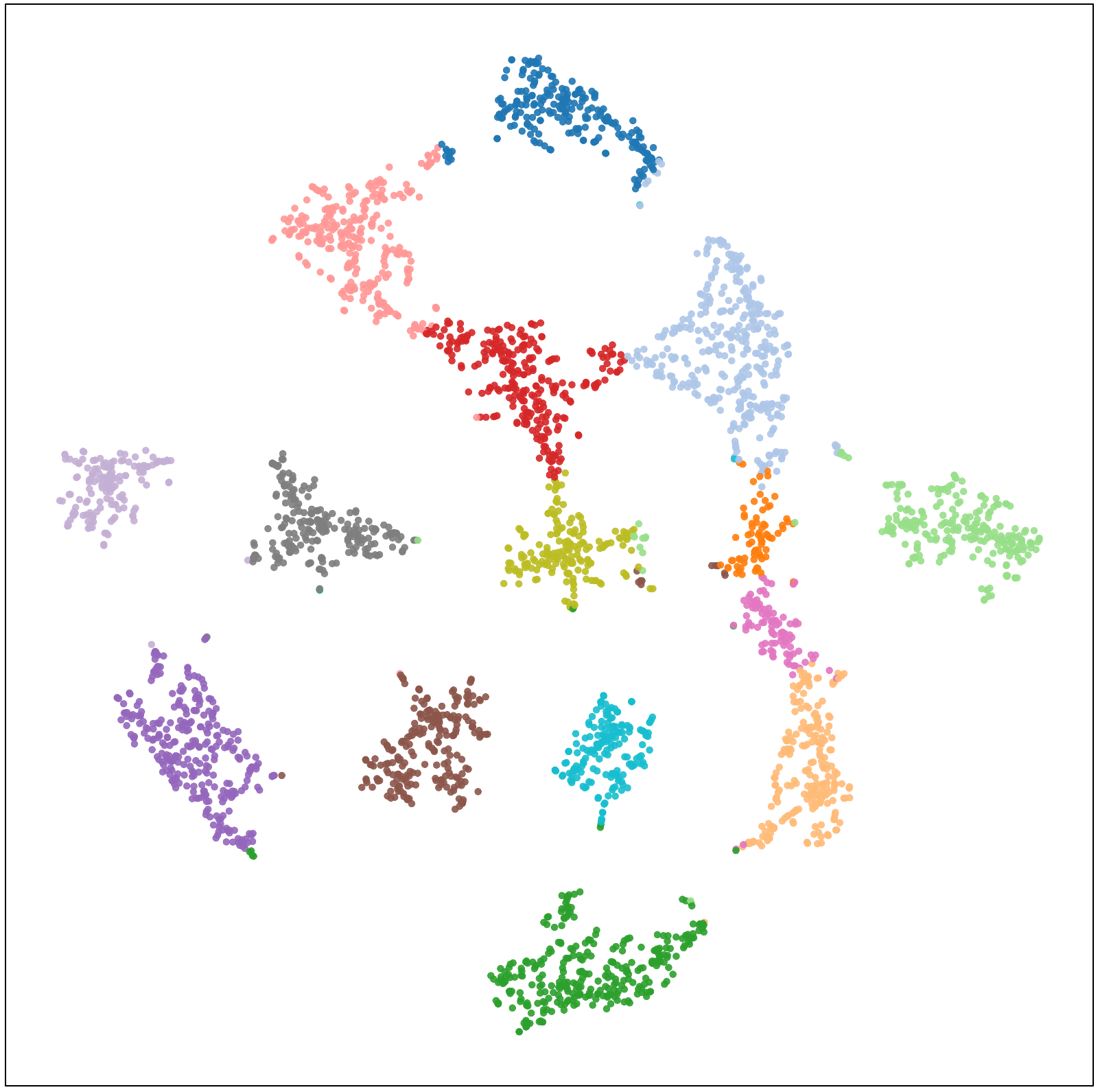}
        \caption{PLCI}
        \label{fig:tsne_ours}
    \end{subfigure}

    \caption{The t-SNE visualization on Scene15 under IIR$=0.5$.}
    \label{fig:tsne_visualization}
\end{figure}

Fig.~\ref{fig:tsne_visualization} shows the t-SNE visualization on Scene15. PLCI yields more compact and better separated clusters than the compared methods. This indicates that posterior-guided latent counterpart inference helps reduce the adverse influence of unavailable and unreliable counterparts, leading to more discriminative representations under imperfect information.

%%%%%%%%%%%%%%%%%%%%%%%%%%%%%%%%%%%%%%%%%%%%%%%%%%%%%%%%%%%%

\newpage
% ~
% \newpage

% \input{checklist.tex}

\end{document}